\documentclass[conference]{IEEEtran}
\IEEEoverridecommandlockouts
\usepackage{cite}
\usepackage{amsmath,amssymb,amsfonts}
\usepackage{algorithmic}
\usepackage{graphicx}
\usepackage{textcomp}
\usepackage{xcolor}
\def\BibTeX{{\rm B\kern-.05em{\sc i\kern-.025em b}\kern-.08em
    T\kern-.1667em\lower.7ex\hbox{E}\kern-.125emX}}

\usepackage{algorithm}
\usepackage{algorithmic}
\usepackage{graphicx}
\usepackage{amsthm}
\usepackage{amsmath}
\usepackage{amssymb}

\newtheorem{definition}{Definition}
\theoremstyle{plain} 
\newtheorem{theorem}{Theorem}

\usepackage{float}
\usepackage{subcaption}
\usepackage{multirow}

\usepackage{booktabs} 

\usepackage{adjustbox}
\usepackage{makecell} 

\usepackage{hyperref}
\usepackage{csquotes}

\usepackage{todonotes}  
\setuptodonotes{inline} 

\begin{document}

\title{Multi-Objective Optimization for Sparse Deep Multi-Task Learning}

\author{\IEEEauthorblockN{Anonymous submission}}

\author{\IEEEauthorblockN{Sedjro S. Hotegni}
\IEEEauthorblockA{\textit{Department of Computer Science} \\
\textit{Paderborn University}\\
Germany \\
sedjro.salomon.hotegni@uni-paderborn.de}
\and
\IEEEauthorblockN{Manuel Berkemeier}
\IEEEauthorblockA{\textit{Department of Computer Science} \\
\textit{Paderborn University}\\
Germany \\
manuelbb@mail.uni-paderborn.de}
\and
\IEEEauthorblockN{Sebastian Peitz}
\IEEEauthorblockA{\textit{Department of Computer Science} \\
\textit{Paderborn University}\\
Germany \\
sebastian.peitz@uni-paderborn.de}
}

\maketitle

\begin{abstract}
  Different conflicting optimization criteria arise naturally in various Deep Learning scenarios. These can address different main tasks (i.e., in the setting of Multi-Task Learning), but also main and secondary tasks such as loss minimization versus sparsity. The usual approach is a simple weighting of the criteria, which formally only works in the convex setting. In this paper, we present a Multi-Objective Optimization algorithm using a modified Weighted Chebyshev scalarization for training Deep Neural Networks (DNNs) with respect to several tasks. By employing this scalarization technique, the algorithm can identify all optimal solutions of the original problem while reducing its complexity to a sequence of single-objective problems. The simplified problems are then solved using an Augmented Lagrangian method, enabling the use of popular optimization techniques such as Adam and Stochastic Gradient Descent, while efficaciously handling constraints. Our work aims to address the (economical and also ecological) sustainability issue of DNN models, with a particular focus on Deep Multi-Task models, which are typically designed with a very large number of weights to perform equally well on multiple tasks. Through experiments conducted on two Machine Learning datasets, we demonstrate the possibility of adaptively sparsifying the model during training without significantly impacting its performance, if we are willing to apply task-specific adaptations to the network weights.
  Code is available at \urlstyle{tt}\url{https://github.com/salomonhotegni/MDMTN}.
\end{abstract}

\begin{IEEEkeywords}
Multi-Objective Optimization, Multi-Task Learning, Model Sparsification, Deep
Neural Networks, Weighted Chebyshev scalarization, Group Ordered Weighted $l_1$
\end{IEEEkeywords}

\section{Introduction}

\textit{Deep Learning} has revolutionized various fields, achieving remarkable success across a wide range of applications. The standard approach to training \textit{Deep Neural Networks} (DNNs) involves minimizing a single loss function, typically the empirical loss, to improve model performance. While effective, this method disregards the potential benefits of simultaneously optimizing multiple objectives, such as reducing model size and computational complexity while maintaining a desirable level of accuracy across different tasks. To address this limitation, one can incorporate a \textit{Multi-Objective Optimization} (MOO) framework into the training process of DNNs, aiming to find a balance between conflicting objectives and enable a more comprehensive exploration of DNN parameterizations. This approach enables the simultaneous optimization of multiple criteria and facilitates the training of DNNs that exhibit optimal trade-offs among these objectives. In the context of Deep Learning, several scenarios require considering trade-offs between objectives that are inherently conflicting. For instance, reducing model size may come at the cost of decreased accuracy, while improving model performance may result in increased model complexity. Neglecting these trade-offs limits the ability to develop versatile and efficient DNN models that excel across various tasks. Rather, Multi-Objective Optimization aims to identify optimal compromises, denoted as the Pareto set, representing a collection of parameterizations of a DNN, each offering a unique trade-off between objectives. By exploring the Pareto set, we gain valuable insights into the interrelationship of different objectives and can make informed decisions regarding model design and parameter selection.

Traditional methods for handling multiple criteria in optimization often involve weighted sum approaches, where objectives are combined into a convex combination. However, this approach assumes convexity and fails to capture non-convex sections of the Pareto front, limiting its effectiveness in complex optimization scenarios \cite{bieker2021treatment, miettinen1999nonlinear}. Moreover, determining appropriate weights for more than two objectives becomes increasingly challenging and subjective. In the context of non-convex problems, we thus need more advanced techniques such as sophisticated \textit{scalarization} techniques \cite{miettinen1999nonlinear}, \textit{gradient-based} approaches \cite{fliege2000steepest}, \textit{evolutionary algorithms} \cite{coello2005evolutionary}, or \textit{set-oriented} techniques \cite{peitz2018gradient, dellnitz2005covering}. In Deep Learning, both evolutionary algorithms and set-oriented methods are too expensive, as they require a very large number of function evaluations. Gradient-based methods, have gained popularity as they offer computational efficiency comparable to single-objective problems. However, these methods typically yield a single solution on the Pareto set, making them unsuitable for scenarios where a decision-maker is interested in multiple possible trade-offs.
To tackle the challenges posed by non-convex problems and conflicting objectives, we propose using a more advanced scalarization approach, namely the \textit{Weighted Chebyshev} (WC) scalarization, coupled with the \textit{Augmented Lagrangian} (AL) method to solve the resulting sequence of scalar optimization problems. Our methodology enables the efficient computation of the entire Pareto set, facilitating a comprehensive exploration of the trade-off environment (i.e. the optimal parameterizations of a DNN model). By considering the reduction of model complexity as an additional objective function, this study additionally aims to address the sustainability issue of DNN models, with a particular focus on Deep Multi-Task models. Specifically, we investigate the feasibility of adaptively sparsifying the model during training to reduce model complexity and gain computational efficiency, without significantly compromising the model's performance. To evaluate the effectiveness of this method, we conduct experiments on two datasets.

The remainder of this paper is organized as follows: In Section 2, we provide a comprehensive review of related works in Multi-Objective Optimization for Deep Neural Network Training. Section 3 presents the methodology, providing a detailed description of a modified WC scalarization method, the AL approach, and the \textit{Group Ordered Weighted $l_1$ (GrOWL)} function, which facilitates the reduction of model size.
Section 4 provides a detailed exposition of the results obtained from two distinct datasets.
Finally, Section 5 concludes the paper.

\section{Related Works}
\label{sec:rel_w}
%
%
\paragraph{Multi-Objective Optimization (MOO)}
Multi-objective optimization is a technique for optimizing several conflicting objectives simultaneously. A \textit{Multi-Objective Optimization Problem} (MOP) involves determining the set of optimal compromises when no specific solution can optimize all objectives simultaneously. To gain a deeper understanding of the Multi-objective optimization problems, we recommend \cite{coello2007evolutionary} and \cite{marler2004survey}. 
Most of the MOO algorithms handle optimization problems using approaches like \textit{evolutionary algorithms} \cite{deb2011multi, konak2006multi}. \textit{scalarization} methods \cite{braun2018scalarized, eichfelder2008adaptive}, \textit{set-oriented} techniques \cite{schutze2013set, peitz2018gradient}, and \textit{gradient-based} approaches \cite{fliege2000steepest}. 
In this work, we focus on scalarization-based algorithms, that allow for preferences in MOO \cite{braun2018scalarized}. Users preferences are integrated by assigning corresponding weights (importance) to each objective. While linear scalarization algorithms struggle to find optimal solutions in non-convex Pareto front cases, in Reinforcement Learning,\cite{van2013scalarized} recommend the Weighted Chebyshev scalarization approach in a Multi-Objective Q-learning framework. This scalarization method uses the Chebyshev distance metric to minimize the distance between the objective functions and a given reference point. 
\cite{banholzer2022trust} solved multi-objective infinite-dimenisonal parameter optimization problems using the \textit{Pascoletti-Serafini} (PS) scalarization approach. As in our work, the obtained scalar optimization problems are solved using the Augmented Lagrangian method.
In \cite{eichfelder2008adaptive}, it has been proved that this PS scalarization approach is equivalent to the WC scalarization, where a variation of the weights and the reference point in WC correspond respectively to a variation of the direction and the reference point in the PS method.
To address the inflexibility of the PS scalarization's constraints, \cite{akbari2018revised} revised a modified PS scalarization algorithm. Nevertheless, for complex MOO problems, the suggested modified PS scalarization algorithm has yet to be tested. 
These methods lead to the identification of the Pareto set, where each Pareto optimal point is obtained under a different choice of scalarization parameter.

\paragraph{Multi-Task Learning (MTL)}
Multi-Task Learning  can be viewed as a form of Multi-Objective Optimization in which the goal is to simultaneously optimize several main tasks 
 \cite{sener2018multi}. In this approach, the DNN is trained to jointly optimize the loss functions of all the tasks, while learning shared representations that capture the common features across the tasks, and task-specific representations that capture the unique features of each task \cite{zhang2021survey}. 
 This optimization technique is known as \textit{hard parameter sharing} \cite{baxter1997bayesian}. 
 Other strategies include \textit{soft parameter sharing} \cite{duong2015low}, 
 \textit{loss construction} \cite{perera2018multi}, and \textit{knowledge distillation} \cite{liu2019improving}.
\cite{sener2018multi} introduced a \textit{Multiple Gradient Descent Algorithm} (MGDA) for Multi-Task Learning.
The suggested method learns shared representations that capture common properties across all the tasks while also optimizing task-specific objectives, using a Deep Neural Network architecture with shared and task-specific layers.
To enable a balanced parameter sharing across tasks, \cite{li2020knowledge} suggested a knowledge distillation-based MTL technique that learns to provide the same features as the single-task networks by minimizing a linear sum of task-specific losses. This technique then relies on the assumption of simple linear relationships between task-specific networks. Moreover, this method requires learning a separate task-specific model for every individual task and subsequently presents a significant challenge, particularly in scenarios involving a substantial number of tasks. 
In robotic grasping, \cite{wong2023fast} proposed \textit{Fast GraspNeXt}, a fast self-attention neural network architecture designed for embedded Multi-Task Learning in computer vision applications. A simple weighted sum of task-specific losses has been used to train the model. 
To enable user-guided decision-making, \cite{mahapatra2021exact} developed an \textit{Exact Pareto optimum} (EPO) search approach for Multi-Task Learning. This method applies gradient-based multi-objective optimization to locate the Pareto optimal solutions. The suggested technique provides scalable descent towards the Pareto front and the user's input preferences within the restrictions imposed. When dealing with disconnected Pareto fronts, EPO search can only fully approximate a single component unless new initial seed points are used. 
\cite{ruchte2021scalable}  recommended using COSMOS, a technique that generates qualitative Pareto front estimations in one optimization phase, instead of identifying one Pareto optimal point at a time. 
Through a linear scalarization with penalty term, the algorithm learns to adapt a single network for all trade-off combinations of the preference vectors.
However, COSMOS cannot handle objectives that do not depend on the input, such as an objective function for sparsity based on model weights, as investigated in our work. 

\paragraph{Model Sparsification (MS)}
Model Sparsification is a technique for reducing the size and computational complexity of Deep Learning models by eliminating redundant parameters and connections. Sparsification aims to provide a more compact model that is computationally efficient without compromising performance considerably \cite{hoefler2021sparsity}. Deep Neural Network sparsification techniques include Weight pruning \cite{molchanov2017variational} and Dropout \cite{ma2021effective}.
\cite{khan2019sparseout} introduced Sparseout, a Dropout extension, that is beneficial for Language models. This approach imposes a $L_q$-norm penalty on network activations. Considering probability as a global criterion to assess weight importance for all layers, \cite{zhou2021effective} proposed a network sparsification technique called probabilistic masking (ProbMask). In ProbMask, the pruning rate is progressively increased to provide a smooth transition from dense to sparse state.
Our study is most related to works that consider model sparsification as a supplementary objective in addition to the primary objective of loss minimization by determining their trade-off. Working with linear classification models, \cite{ustun2016supersparse} presented SLIM, a data-driven scoring system constructed by solving an integer program that directly encodes measurements of accuracy and sparsity. The sparsity is managed through the $l_0$-norm of the model parameters added to the $0$-$1$ loss function (for accuracy) and a small $l_1$-penalty (to limit coefficients to coprime values). In \cite{zhou2016less}, Group-Lasso has been used in training \textit{Convolutional Neural Networks} (CNNs) as sparse constraints imposed on neurons during the training stage. This method may, however, fail in the presence of highly correlated parameters. To simultaneously prune redundant neurons and tying parameters associated with strongly correlated parameters, \cite{zhang2018learning} used the \textit{Group Ordered Weighted $l_1$} (GrOWL) regularizer for Deep Neural Networks, with a low accuracy loss. GrOWL, promotes sparsity while learning which sets of parameters should have a common value.
In our case, we suggest an appropriate Multi-Objective Optimization approach for training a DNN in terms of its primary objective functions, as well as the GrOWL function as a secondary objective function.

\section{Methodology}
\label{sec:meth}
\subsection{Multi-Objective Optimization}
Consider a Deep Neural Network model that is meant to accomplish $m \in \mathbb{N}$ primary tasks simultaneously. Let $\mathcal{L}_i\ (i=1,\dots,m)$ be the loss functions associated with each task, and $\mathcal{L}_0$ be an extra objective function whose minimization leads to a sparse model. Our goal is to train the DNN with respect to the criteria $\mathcal{L}_i\ (i=0,\dots,m)$. This calls for the discovery of a set of solutions $x^*$ to the following problem.
\begin{align*}
        \min_{x \in \Omega} \left( \begin{array}{c}
            \mathcal{L}_0(x)  \\
            \mathcal{L}_1(x)  \\
            \vdots \\
            \mathcal{L}_m(x)
        \end{array} \right),
        \label{eq:MOO}
        \tag{MOP}
\end{align*}
where $\Omega$ is the feasible set of the decision variable $x$ that represents the model parameterization (i.e., the network weights), and $\mathcal{L}_i\ (i = 0,\dots,m)$ are continuously differentiable. In the scenario where $m = 1$, we transform a standard learning problem with a single task and sparsity constraints into an MOP, while for $m > 1$, the original problem inherently constitutes a Multi-Task Learning problem. In MOO, we look for Pareto optimal solutions \cite{pareto1964cours, deb2005searching}. 
\begin{definition}
Let $x^*$ be a solution of a (MOP).
\begin{itemize}
        \item $x^*$ is said to be \textbf{Pareto Optimal} with respect to $\Omega$, if there is no other solution that dominates it. This means that there is no $x' \in \Omega$ such that:
    \begin{itemize}
        \item $\forall\ i = 0, \dots, m,\ \mathcal{L}_i(x')\le \mathcal{L}_i(x^*)$
        \item $\exists\ j\in \{0, \dots, m\}\ |\ \mathcal{L}_j(x') < \mathcal{L}_j(x^*)$
    \end{itemize}
        \item $x^*$ is said to be \textbf{weakly Pareto Optimal} with respect to $\Omega$, if there is no other solution $x' \in \Omega$ such that:
        
$\forall\ i = 0, \dots, m,\ \mathcal{L}_i(x') < \mathcal{L}_i(x^*)$
\end{itemize}
\end{definition}

The Pareto optimal solutions provide the best achievable trade-offs between the objectives, as enhancing one would necessitate sacrificing another. Solving the MOP involves exploring the trade-off space to identify the \textbf{\textit{Pareto set}} (set of Pareto optimal solutions), represented as the \textbf{\textit{Pareto front}} in the objective space.

By directly solving the MOP, the Pareto set can be discovered entirely at once or iteratively. Scalarization-based methods are designed to provide one optimal solution at a time within the Pareto set by addressing a related problem. They \enquote{simplify} the MOP to a scalar optimization problem by assigning a weight/importance to each objective function \cite{eichfelder2008adaptive}. This is referred to as a user guided decision-making process, since it results in a solution that is aligned with the user's demands (a predefined importance vector, for example). Finding various solutions within the Pareto set necessitates solving the new problem several times while changing certain parameters (usually, the importance vector). 
Weighted Chebyshev scalarization is one of the most effective
scalarization methods for Multi-Objective Optimization
due to its ability to find every point of the Pareto front
by variation of its parameters.

\subsection{Weighted Chebyshev Scalarization}
Assume that $k = (k_0, \dots, k_m)$ is a predefined importance vector and $a$ is a specified reference point, such that
$\forall i\in\{0,\dots,m\},\ k_i \ge 0$ with $\sum_{i=0}^{m}k_i = 1$ and:
$\forall x\in \Omega\ \text{,}\ a_i < \mathcal{L}_i(x).$
The modified \textit{Weighted Chebyshev} scalarization approach $\big(\text{Modified WC}(k,a)\big)$, suggested in~\cite{kaliszewski2012quantitative}, is defined by:

\begin{align*}
\begin{array}{rl}
    \text{Min} &\ t\\
    \text{s.t.}: &\ k_i\bigg[\big(\mathcal{L}_i(x)-a_i\big) + \epsilon\sum_{i=0}^{m}\big(\mathcal{L}_i(x)-a_i\big)\bigg] - t\le 0,\\
    \ &\ t\in \mathbb{R},\ x\in \Omega\\
    \ &\ i \in \{0, \dots, m\}
\end{array},
\label{eq:mod_WCS}
\tag{Modified WC(k,a)}
\end{align*}
where $\epsilon$ is a sufficiently small positive scalar, constant for all objectives, and can be interpreted as a permissible disturbance of the preference cone $\mathbb{R}_+^{m+1}$ \cite{kaliszewski1987modified}.
\cite{kaliszewski2012quantitative} demonstrated that solving problem \eqref{eq:mod_WCS} always yields Pareto optimal solutions (proper Pareto Optimality).

Therefore, finding all the Pareto Optimal solutions requires solving problem \eqref{eq:mod_WCS} with different importance vectors $k$, while keeping the reference point $a$ fixed \cite{eichfelder2008adaptive}. We apply the Augmented Lagrangian transformation to solve this problem while carefully dealing with its constraints. 

\subsection{Augmented Lagrangian Transformation}

The Augmented Lagrangian (AL) approach is well-known for its efficacy in tackling constrained optimization problems. It was suggested by Hestenes and Powell \cite{hestenes1969multiplier, rockafellar1973multiplier} and has since become a popular optimization approach due to its robustness and adaptability. We transform problem \eqref{eq:mod_WCS} using the AL method and the following new loss function is then minimized with respect to $(x,t,\lambda)$:
\begin{equation*}
L_A(x, t, \lambda,\mu) = t + \lambda^TH(x,t) + \frac{1}{2}\mu \big|\big|H(x,t)\big|\big|^2_2
\label{eq:func_AL}
\tag{1}
\end{equation*}

where $\mu>0$ is the augmentation term coefficient,\\ $\lambda = (\lambda_0, ..., \lambda_m)\in\mathbb{R}^{m+1}$ and $H(x,t)$ is the vector of inequality constraints.


The transformations made from \eqref{eq:MOO} to \eqref{eq:func_AL} allow the DNN to be trained using popular optimizers such as Adam and Stochastic Gradient Descent (SGD) while solving the MOP. Our method necessitates a careful selection of the objective function $\mathcal{L}_0$ in order to boost the model's sparsity. For this purpose, we consider the Group Ordered Weighted $l_1$ (GrOWL) function that was used in \cite{zhang2018learning} as a regularizer for learning DNNs. 

\subsection{Group Ordered Weighted $l_1$ (GrOWL)}
\label{sec:growl}
The Group Ordered Weighted $l_1$ (GrOWL) is a sparsity-promoting regularizer that is learned adaptively throughout the training phase, and in which groups of parameters are forced to share the same values.
Consider a feed-forward DNN with $L$ layers. Let $N_l$ denotes the number of neurons in the $l$-th layer, $W_l \in \mathbb{R}^{N_{l-1}\times N_l}$ the weight matrix of each layer, and $w_{[i]}^{(l)}$ the row of $W_l$ with the $i$-th largest $l_2$-norm. The layer-wise GrOWL regularizer \cite{zhang2018learning} is given by:

\begin{equation*} 
\mathcal{G}_{\theta^{(l)}}(W_l) = \sum_{i=1}^{N_{l-1}}\theta_i^{(l)}||w_{[i]}^{(l)}||_2
\label{eq:layer_GrOWL}
\tag{2}
\end{equation*} where $\theta^{(l)} \in \mathbb{R}_{+}^{N_{l-1}}$, $\theta_1^{(l)}> 0$, and $\theta_1^{(l)} \ge \theta_2^{(l)} \ge \dotsm \ge \theta_{N_{l-1}}^{(l)} \ge  0.$ 
In this paper, we consider the GrOWL-Spike weight pattern\cite{oswal2016representational}, which prioritizes the largest term while weighting the rest equally:

\begin{align*}
    \begin{cases}
      \theta_1 = \beta_1 + \beta_2\\
      \theta_i = \beta_2,\ \text{for $i = 2, ..., N_{l-1}$ } \\
      \beta_1,\ \beta_2 >0.
    \end{cases}       
\label{eq:GrOWL_spike}
\tag{GrOWL-Spike}
\end{align*} 

We define the objective function $\mathcal{L}_0$ as a sum of the layer-wise GrOWL penalties:

\begin{align*}
\mathcal{L}_0(x) = \sum_{i=1}^{L} \mathcal{G}_{\theta^{(l)}}(W_l)
\label{eq:GrOWL_l0}
\tag{3}
\end{align*}

In the case of convolutional layers, each weight $W_l \in \mathbb{R}^{F_w\times F_h\times N_{l-1}\times N_l}$ (where $F_w$ is the filter width, and $F_h$ the filter height) is first reshaped into $W_l^{2D} \in \mathbb{R}^{N_{l-1}\times(F_wF_hN_l)}$ before applying the GrOWL regularizer.
Applying GrOWL to a layer during training allows for removing insignificant neurons from the previous layer, while encouraging similarity among the highly correlated ones.


There are two major stages during the training process. The first is a sparsity-inducing and parameter-tying phase, in which we seek for an appropriate sparse network from the original DNN by identifying irrelevant neurons that should be eliminated and clusters of strongly correlated neurons that must be encouraged to share identical outputs. At the end of each epoch, a proximity operator 
$\operatorname{prox}_{\mathcal{G}_{\theta}}(.)$ is applied to the weight matrices for efficient optimization of \eqref{eq:GrOWL_l0} (for details, see \cite{zhang2018learning}).
The DNN is then retrained (second phase) exclusively in terms of the identified connectivities while promoting parameter sharing within each cluster. 

\subsection{Deep Multi-Task Learning Model}
\label{sec:Dmoo}
\begin{figure}[H]
  \centering
  \includegraphics[width=0.5\textwidth]{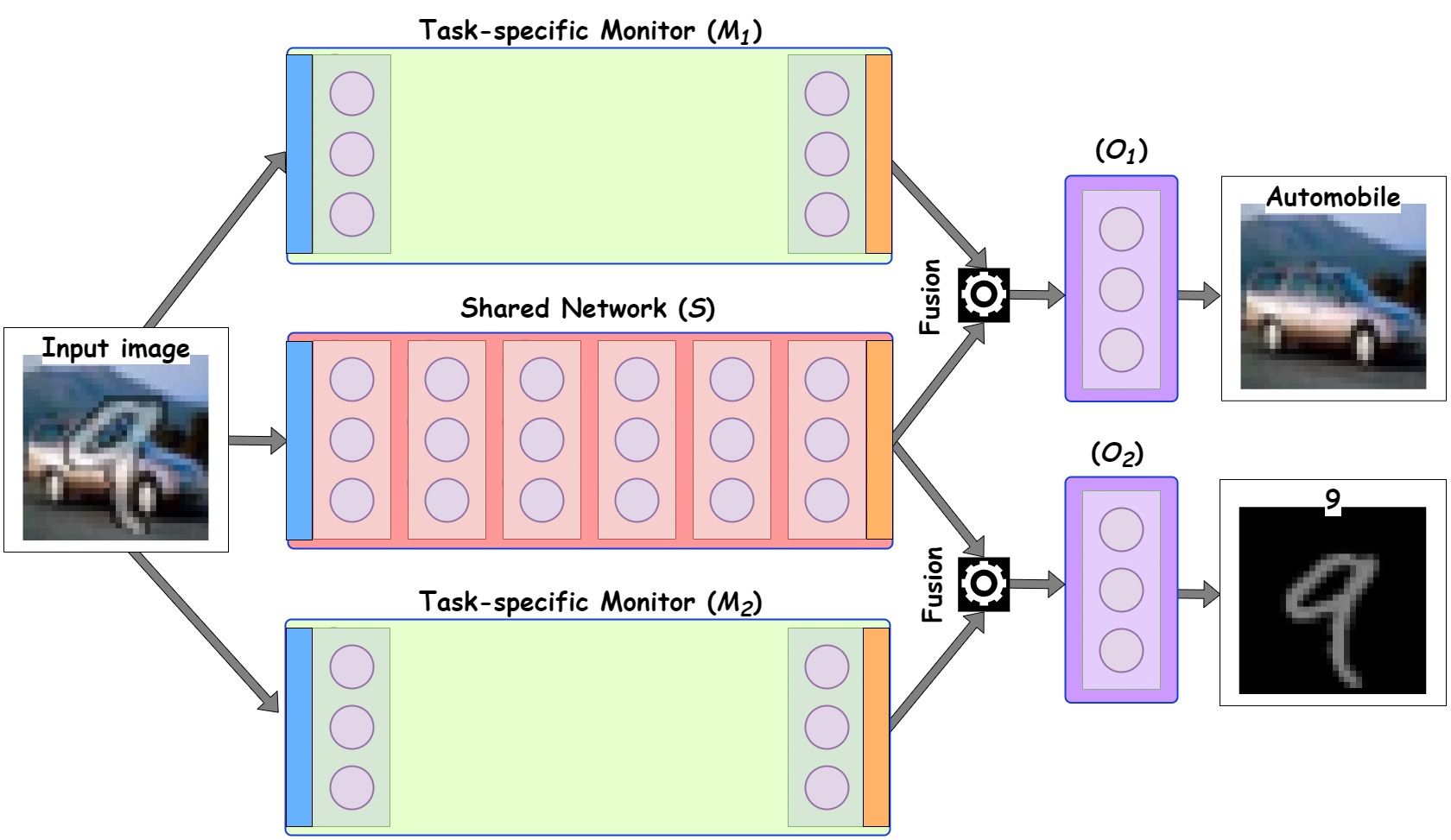}
  \caption{Diagram of the Monitored Deep Multi-Task Network (MDMTN) with 2 main tasks.}
  \label{fig:fig1}
\end{figure}

A Multi-Task Learning model is required when addressing more than one main task $(m>1)$. In the most frequently used architecture, \textit{hard parameter sharing}, the hidden layers are completely shared across all tasks, while maintaining multiple task-specific output layers \cite{sener2018multi}. However, this method is only effective when the tasks are closely associated. Other alternatives have been proposed, including Multi-Task Attention Network (MTAN) \cite{liu2019end}, Knowledge Distillation for Multi-Task Learning (KDMTL) \cite{li2020knowledge}, and Deep Safe Multi-Task Learning (DSMTL) \cite{yue2021deep}.
These approaches are based on the idea of allowing each task to have its fair opportunity to contribute, while simultaneously learning shared features.
However, as a \textit{soft parameter sharing} approach, scalability is an issue with MTAN, since the network's complexity expands linearly with the number of tasks \cite{vandenhende2021multi}. On the other hand, KDMTL and DSMTL involve duplicating the shared network for every single task. In this paper, we present a novel Deep MTL model in which the task-specific information that the shared network may miss are learned using task-specific monitors. These monitors are sufficiently simple to ensure that they do not considerably increase the model's complexity while dealing with many tasks. Our proposed model, namely the \textit{Monitored Deep Multi-Task Network} (MDMTN), is shown in Figure \ref{fig:fig1}. 

The Task-specific monitors ($\mathcal{M}_i$) can only have up to two layers and are designed to have the same input and output dimensions as the Shared network ($\mathcal{S}$) to enable fusion. Given an input $u$, each Task-specific output network ($\mathcal{O}_i$) then produces
\resizebox{0.4\linewidth}{!}{$\mathcal{O}_i\big(\alpha_1^{(i)}\mathcal{S}(u)+\alpha_2^{(i)}\mathcal{M}_i(u) \big),$} where $\alpha_1^{(i)}$ and $\alpha_2^{(i)}$ are real values learned by the model.
Regarding the secondary objective (sparsity), the Group Ordered Weighted $l_1$ (GrOWL) is applied to all layers except the input layers (to preserve data information) and the Task-specific output layers (for meaningful outputs).
No neurons are eliminated from these layers at the end of the training stage. 
Algorithm \ref{alg:Train} presents the training process.

\begin{algorithm}[H]
\scriptsize
    \caption{Training}
    \label{alg:Train}
    \textbf{Input}: Consider a DNN model that is meant to accomplish $m\ (m \in \mathbb{N})$ primary tasks simultaneously. Having $k = (k_0, \dots, k_m)$, fix $a$ w.r.t. the loss functions $f_0(x), \dots, f_m(x)$. 
    Choose a maximum sparsity rate $s_{l_{max}}$ of a layer, and a minimum sparsity rate $s_{min}$ of a model.\\ 
    Choose an optimizer (Adam, SGD, \dots). Let $\Omega$ denotes the set of all parameterizations $(x, t)$ of the model.\\
    \textbf{Output}: A trained, sparse and compressed DNN.\\
	\begin{algorithmic}[1]
            \STATE{\bf PHASE 1:} Sparsity-inducing and parameter-tying.
            \STATE{\bf Initialize} the Lagrange multipliers $\lambda$ and $\mu$. 
    		\FOR{$j=1$ to $M_1$ iterations}
                    \STATE{\bf Set} the maximum sparsity rate of a layer to\\ $s_{l_{max}}^j = j\big(\frac{s_{l_{max}}}{M_1}\big)$\\
                    \FOR{$i=1$ to $p$ epochs}
            		\STATE{\bf optimize} the loss function \eqref{eq:func_AL} to update the model parameters.\\
                   \STATE{\bf GrOWL:} Remove insignificant neurons to obtain a model with a sparsity rate $s_j$ and save the model if $s_j\ge s_{min}$ with better performance on the main tasks.
    	        \ENDFOR
    		      \STATE{\bf update} the Lagrange multipliers $\lambda$ and $\mu$
                \ENDFOR
            \STATE{\bf GrOWL:} Identify the clusters of neurons that should be encouraged to share the same values.
            \STATE{}
            \STATE{\bf PHASE 2:} Parameter sharing (Retraining)
            \STATE{} Maintain the previously identified sparse model architecture and, for $M_2$ maximum iterations, repeat steps $2-10$ except $7$, which should be replaced by $15$:
            \STATE{\bf GrOWL:} Force parameter sharing within the identified clusters at $11$.
	\end{algorithmic} 
    \end{algorithm}

Considering a 2D (reshaped) weight matrix $W_l$ of the $l^{th}$ layer, the rows represent the outputs from the  $(l-1)^{th}$ layer's neurons. Thus, given a threshold $\tau$, the non-significant rows (in terms of norm) are set to zero at each epoch during the first stage of the training phase, and the model is saved if the average accuracy on the primary tasks improves. For a model to be saved, we define a minimal sparsity rate $s_{min}$. We set a maximum sparsity rate $s_{l_{max}}$ for each layer to avoid irrelevant layers (or zero layers).
At the end of this stage, we obtain a sparse model, ignoring the connectivities involving the neurons associated to the zero rows of the weight matrices. For the identification of the clusters of neurons that should be forced to produce identical outputs during the second stage of the training phase, we follow \cite{zhang2018learning} and use the \textit{Affinity Propagation} method based on the following pairwise similarity metric:

\begin{align*}
\mathcal{S}_l(i,j) = \dfrac{W_{l,i}^TW_{l,j}}{max\big(||W_{l,i}||_2^2, ||W_{l,j}||_2^2)} \in [-1, 1]
\label{eq:GrOWL}
\end{align*}
with $W_{l,i}$ and $W_{l,j}$, respectively, the $i^{th}$ and $j^{th}$ rows of $W_l$.

In the subsequent phase, the most recent model obtained in the initial stage advances along with the clustering information of its weight matrices. This model undergoes a retraining stage, focusing solely on the interconnections among its important neurons, and at each epoch, for each weight matrix $W_l$, the rows belonging to the same cluster are substituted with their mean values. As a result, the final model is a compressed and sparse version of the original one.


\cite{sener2018multi}'s approach demonstrates a lack of tradeoffs between tasks, leading to the collapse of the Pareto front into a single point, thereby eliminating task conflicts from an optimization perspective. However, due to the rising concern over the carbon footprint of AI \cite{dodge2022measuring}, there is a growing interest in developing smaller models that are better tailored to specific tasks. Therefore, in the present study, we address the following inquiry for a given MTL problem: At what level of sparsity or network size do tasks genuinely exhibit conflicts, resulting in a true tradeoff?

\section{Experiments}
\label{sec:Exp}
\begin{figure}[H]
    \centering
    \includegraphics[width=0.4\textwidth]{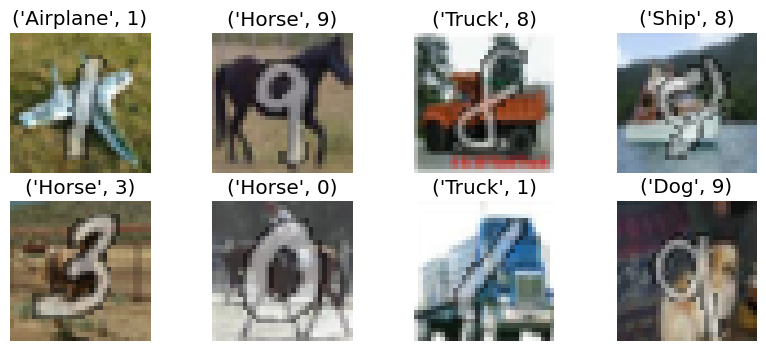}
    \caption{Sample images from the Cifar10Mnist dataset}
    \label{fig:sample_CMdata}
\end{figure}

We start with the MultiMNIST dataset\cite{sabour2017dynamic}, derived from the MNIST dataset of handwritten digits \cite{lecun2010mnist} by gluing two images together, with some overlap.
The primary tasks are the classification of the top left and bottom right digits of each image. To study a more complicated problem with less aligned tasks, we have created our own dataset, namely the Cifar10Mnist dataset (Figure \ref{fig:sample_CMdata}), which contains padded MNIST digits on top of the CIFAR-10 images \cite{Krizhevsky09learningmultiple}.

On this new dataset, we investigate two primary objectives: identifying the background CIFAR-10 image and classifying the padded MNIST digit of each image. Since we are dealing here with two different data sources (CIFAR-10 and MNIST), that have distinct characteristics and representations, it is challenging to simultaneously learn meaningful features and representations for both tasks. The MultiMNIST dataset, on the other hand, typically involves only one data source, making it less complex in terms of data diversity. 
These two types of MTL datasets then allows us to generalize our approach more effectively. 
\cite{sener2018multi}'s approach (MGDA) and the \textit{Hard Parameter Sharing} (HPS) model architecture are used as a baseline algorithm and a baseline MTL model.  The metrics used to evaluate the model architectures are: Sparsity Rate (SR) = $\frac{\# zero\_ params}{\# total\_ params}$, Parameter Sharing (PS) = $\frac{\# nonzero\_ params}{\# unique\_ params}$, and Compression Rate (CR) = $\frac{\# total\_ params}{\# unique\_ params}$.

\subsection{Sparsification Effects on Model Performance}
\label{sec:sparsitySTUDY}


\begin{figure*}[t]  
        \begin{table}[H]
          \begin{adjustbox}{width=2\columnwidth}
          \centering
          \begin{tabular}{cccccccc}
            \Xhline{1.5pt} 
            \multirow{2}{*}{Dataset} &\multirow{2}{*}{Architecture} &\multirow{2}{*}{Method} & \multirow{2}{*}{(SR; CR; PS)} & \multicolumn{3}{c}{Accuracy} &Training time \\
            \cline{5-7}
            & & & & Task 1 & Task 2 & (average) & (minutes) \\
            \Xhline{1.5pt} 
            \multirow{9}{*}{\rotatebox[origin=c]{90}{MultiMNIST}} &\multirow{3}{*}{HPS} & MGDA &  $(0\%;1;1)$ & \fbox{$97.26\%$} & $95.90\%$ & \fbox{$96.580\%$} & $17$ \\
            & & Ours: $k = (0.0, 0.4, 0.6)$ & $(0\%;1;1)$ & $96.74\%$ & $95.53\%$ &  $96.134\%$ & $5$\\
            & & Ours: $k = (10^{-4}, 0.5, 0.4999)$ & \fbox{$(11.83\%; 1.13; 1)$} & $96.78\%$ & \fbox{$96.14\%$} &  $96.46\%$ & \fbox{\textbf{4}}\\
            \cline{2-8}
            &\multirow{3}{*}{MDMTN (Ours)} & MGDA &  $(0\%;1;1)$ & $96.88\%$ & \fbox{\textbf{96.63\%}} & $96.755\%$ & $41$ \\
            & & Ours: $k = (0, 0.35, 0.65)$ & $(0\%;1;1)$ & $97.10\%$ & $96.39\%$ & $96.744\%$ & $12$\\
            & & Ours: $k = (10^{-3}, 0.2, 0.799)$ &  \fbox{\textbf{(57.26\%;2.7;1.15)}} &  \fbox{\textbf{97.29\%}} &  $96.40\%$ & \fbox{\textbf{96.845\%}} & \fbox{$9$} \\
            \cline{2-8}
            & \multicolumn{2}{c}{KDMTL} &  $(0\%;1;1)$ & $96.16\%$ & $95.56\%$ & $95.86\%$ & $19$ \\
            & \multicolumn{2}{c}{MTAN} &  $(0\%;1;1)$ & $96.40\%$ & $95.27\%$ & $95.835\%$ & $23$ \\
            & \multicolumn{2}{c}{STL} & $(0\%;1;1)$ & $97.23\%$ & $95.90\%$ & $96.565\%$ & $16$ \\
            \Xhline{1.5pt} 
            \multirow{9}{*}{\rotatebox[origin=c]{90}{Cifar10Mnist}} & \multirow{3}{*}{HPS} & MGDA &  $(0\%;1;1)$ & $45.75\%$ & \fbox{$96.78\%$} & $71.265\%$ & $9$ \\
            & & Ours: $k = (0.0, 0.5, 0.5)$ & $(0\%;1;1)$ & $54.00\%$ & $95.14\%$ &  $74.570\%$ & $6$ \\
            & & Ours: $k = (10^{-1}, 0.6, 0.3)$ & \fbox{$(13.03; 1.25; 1.10)$} & \fbox{$56.64\%$} & $94.81\%$ &  \fbox{$75.724\%$} & \fbox{\textbf{5}} \\
            \cline{2-8}
            & \multirow{3}{*}{MDMTN (Ours)} & MGDA &  $(0\%;1;1)$ & $58.17\%$ & \fbox{97.04\%} & $77.605\%$ & $23$ \\
            & & Ours: $k = (0, 0.55, 0.45)$ & $(0\%;1;1)$ & $59.25\%$ & $95.19\%$ & $77.22\%$ & $22$\\
            & & Ours: $k = (10^{-2}, 0.8, 0.19)$ &  \fbox{\textbf{(29.31\%;1.94;1.37)}} &  \fbox{\textbf{61.05\%}} &  $94.26\%$ &  \fbox{\textbf{77.654\%}} & \fbox{$19$} \\
            \cline{2-8}
            & \multicolumn{2}{c}{KDMTL} &  $(0\%;1;1)$ & $58.47\%$ & $96.18\%$ & $77.325\%$ & $23$ \\
            & \multicolumn{2}{c}{MTAN} &  $(0\%;1;1)$ & $52.91\%$ & $95.58\%$ & $74.245\%$ & $25$ \\
            & \multicolumn{2}{c}{STL} & $(0\%;1;1)$ & $58.11\%$ & \textbf{97.07\%} & $77.59\%$ & $23$ \\
            \Xhline{1.5pt} 
          \end{tabular}
           \label{table:resultsMM}
           \end{adjustbox}
        \end{table}
    \captionof{table}{Results on the MultiMNIST and Cifar10Mnist data. Tasks 1 and 2 represent the classification of left and right digits for MultiMNIST and CIFAR-10 and MNIST images for Cifar10Mnist, respectively. For each metric, the best performance per architecture is boxed, while the best overall performance is in bold.  Higher SR and CR are preferable, while PS $> 1$ is better.}
\end{figure*}


\begin{figure*}[t]
  \begin{subfigure}[t]{0.6\textwidth}
    \centering
        \includegraphics[width=1\textwidth]{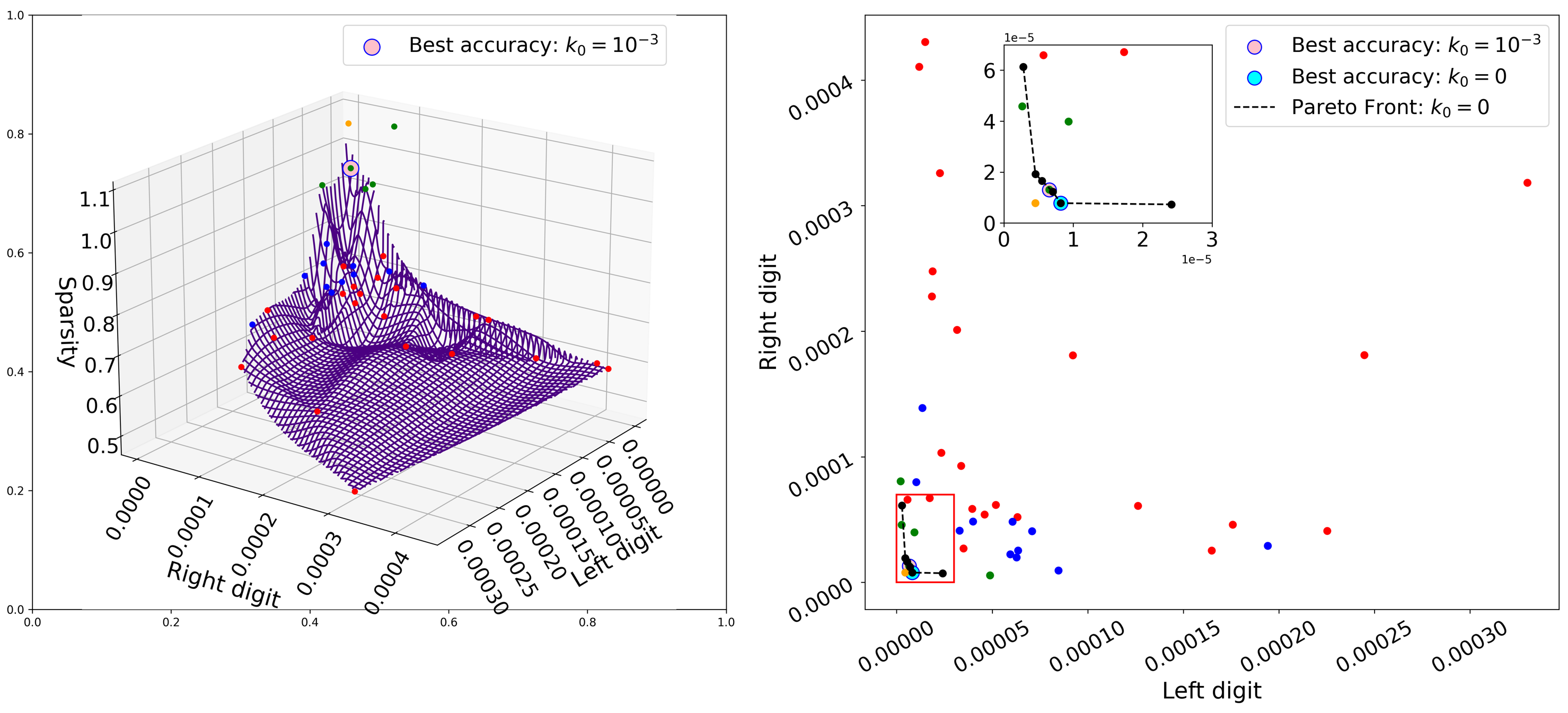}
    \end{subfigure}%
    \hspace{0.01\textwidth}
\begin{subfigure}[t]{0.34\textwidth}
    \centering
        \includegraphics[width=0.9\textwidth]{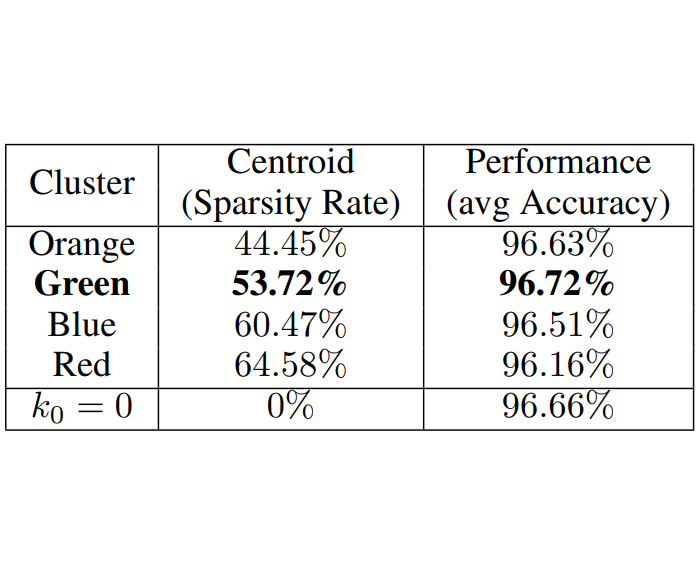}
    \end{subfigure}
    \caption{(MultiMNIST): 3D and 2D visualizations of $39$ Pareto optimal points $(k_0 \neq 0)$ obtained with different preference vectors $k$ (considering the loss functions), and clustered by sparsity rate, as specified in the table on the right.
    We add the Pareto front obtained without taking into account the sparsity objective to the 2D projection $(k_0 = 0,\ \text{in black})$.
    }
    \label{fig:sparsitySTUDY_MM}
\end{figure*}

To investigate the impact of the secondary objective (sparsification) on the primary objectives, we examine the 3D Pareto front obtained with all objective functions and its 2D projection onto the primary objectives space by evaluating 90 different preference vectors $k = (k_0, k_1, k_2)$ and a fixed reference point $a = (0, 0, 0)$. 
The sparsity coefficient $k_0$ values used are $0, 10^{-1}, 10^{-2}, 10^{-3}$ and $10^{-4}$. 
Using this approach, we obtain $17$ to $19$ points for each value of $k_0$, i.e., $90$ points in total. Due to stochasticity during training, we had to perform a consecutive $\epsilon$-nondominance test \cite{mca23020030}, to filter out the points that are non-dominated (within a tolerance of $\epsilon$).
The Shared Network module in all MTL architectures used is consistent across each dataset.
We study the Pareto set and identify the single point with the best overall performance when averaging over all the main tasks. Given the exceptionally high dimensions of MTL models, MOEA/D \cite{zhang2007moea}, as an evolutionary algorithm, struggles to yield meaningful results on the primary tasks and is consequently omitted from our report.

\paragraph{MultiMNIST dataset:}
For $\epsilon = 0$ (exact non-dominance), we get $39$ Pareto optimal points on MultiMNIST (case $k_0 \neq 0$). The sparse model obtained with $k = (10^{-3}, 0.2, 0.799)$ produces the best average accuracy of $96.845\%$ on the main tasks, despite having removed more than half $(SR = 57.26\%)$ of the original model's parameters (neurons).  
As shown in Table 1, this model promotes parameter tying $(PS = 1.15>1)$ and compresses the original model by a factor of more than two $(CR = 2.7)$. The results also show that for each method, the overall performance of our proposed MDMTN model architecture exceeds that of the HPS architecture. Furthermore, the sparse MDMTN model obtained through our approach surpasses the performance of individually learning each task (\textit{Single-Task Learning - STL}), as well as that of the MGDA, KDMTL, and MTAN methods. Figure \ref{fig:sparsitySTUDY_MM} shows the visualizations of the obtained Pareto optimal points on MultiMNIST, clustered by sparsity rate. 
An approximately convex 3D Pareto front is obtained.
Thus, when moving along the front from one solution to another, improvements in one objective are achieved at the expense of others in a generally smooth and consistent way. 
We project this Pareto front onto the space of the main objective functions to better understand the influence of the secondary objective on model performance, and according to the clustering information (right table), sparse models with a sparsity rate of roughly $53\%$ increase model performance while greatly reducing its complexity. Therefore, although a largely sparse model struggles to jointly optimize the primary objectives in comparison to the original model, given a preference vector $k$, our technique may adaptively identify a suitable sparse model from the original one at an acceptable cost. Moreover, the result emphasizes that an even larger sparsity can be achieved, but at the cost of having to trade between the two main tasks considerably. However, considering the goal of reducing the size and the training effort of large-scale ML models, this might be very useful in practical situations, where we can adjust the weights of a small network to adapt to different tasks.

\paragraph{Cifar10Mnist dataset:}
As expected, a significant gap in model sparsity leads to poor performance on the Cifar10Mnist dataset, since the two tasks are highly distinct and demand a somewhat robust model to learn their shared representations effectively. Therefore, during the first stage of the training phase, we consider $s_{l_{max}} = 30\%$ (maximum sparsity rate of a layer).
For $\epsilon = 0$ (exact non-dominance), we get $35$ Pareto optimal points (case $k_0 \neq 0$), and obtain the best model performance of $77.65\%$ with $k = (10^{-2}, 0.8, 0.19) $. This results in a sparse model that ignores $29.31\%$ of the original model's parameters and encourages parameter sharing $(PS = 1.37 > 1)$. Its compression rate is $CR = $1.94. Table 1 shows that MGDA sacrifices more performance on Task 1 (CIFAR-10) in favor of Task 2 (MNIST). Although it excels in MNIST digit classification, this advantage is achieved at the expense of overall performance, unlike our method (with sparsity), which exhibits superior performance across the tasks. Similar to the scenario with MultiMNIST, the results obtained on Cifar10Mnist demonstrate that, across all methods, our proposed MDMTN model architecture consistently outperforms the HPS, KDMTL, and MTAN architectures. Notably, the performance improvement is even more significant in this case.

\begin{figure}[H]
    \centering
    \includegraphics[width=0.28\textwidth]{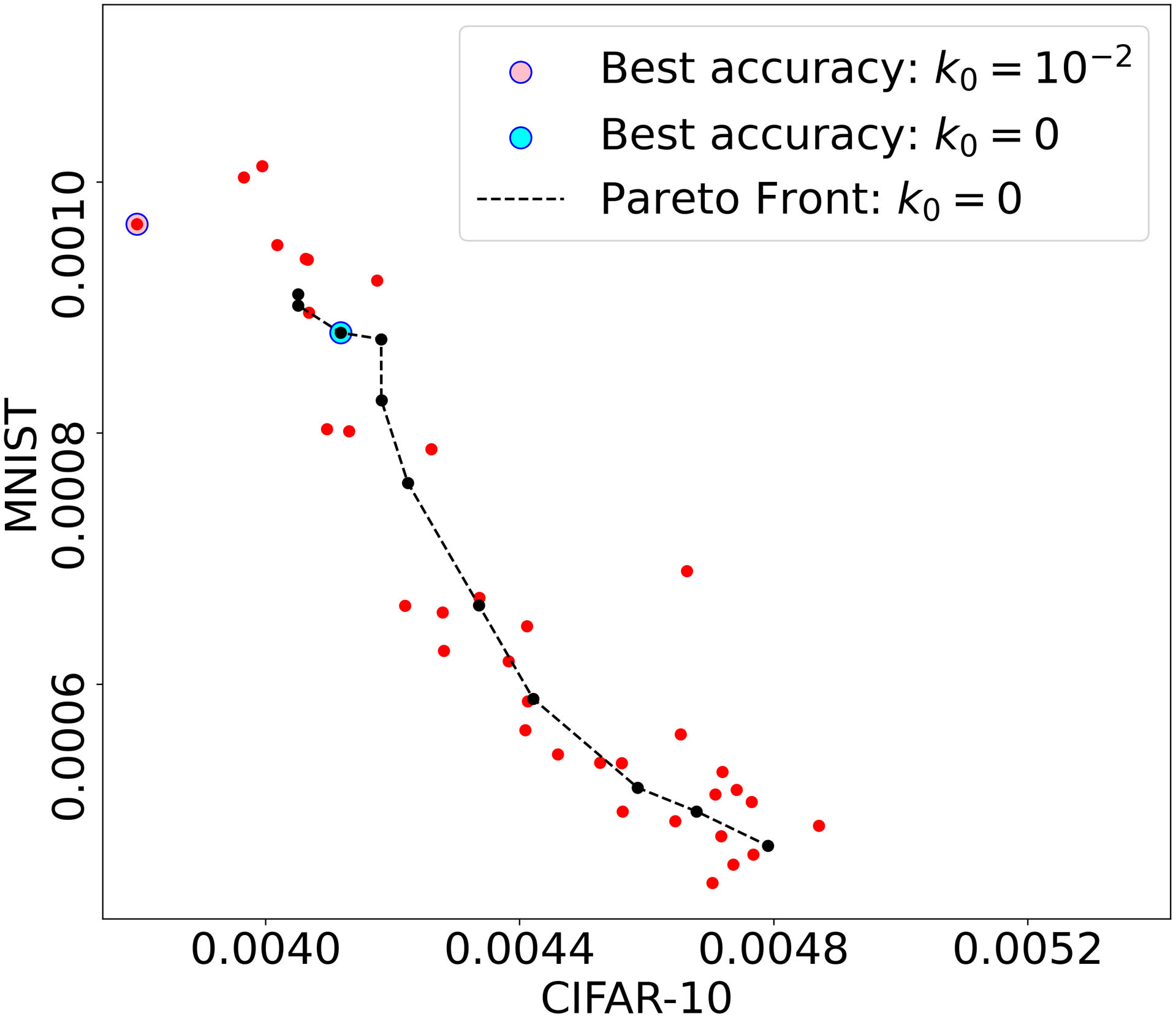}
    \caption{(Cifar10Mnist): 2D projection of $35$ Pareto optimal points ($k_0 \neq 0$, in red) and the Pareto front obtained without taking into account the sparsity objective ($k_0 = 0$, in black).}
    \label{fig:sparsitySTUDY_CM}
\end{figure}

Since the model sparsity tolerance range is small, we do not study a 3D Pareto front for this problem. We show the 2D projection of the obtained Pareto optimal points in Figure \ref{fig:sparsitySTUDY_CM}.
Each of the $35$ preference vectors produces a model with a sparsity rate of about $29\%$, and their performances vary between $74.46\%$ and $77.65\%$.

\begin{figure}[H]
\centering
  \begin{subfigure}[t]{0.245\textwidth}
        \caption{MDMTN on MultiMNIST}
        \includegraphics[width=1.\textwidth]{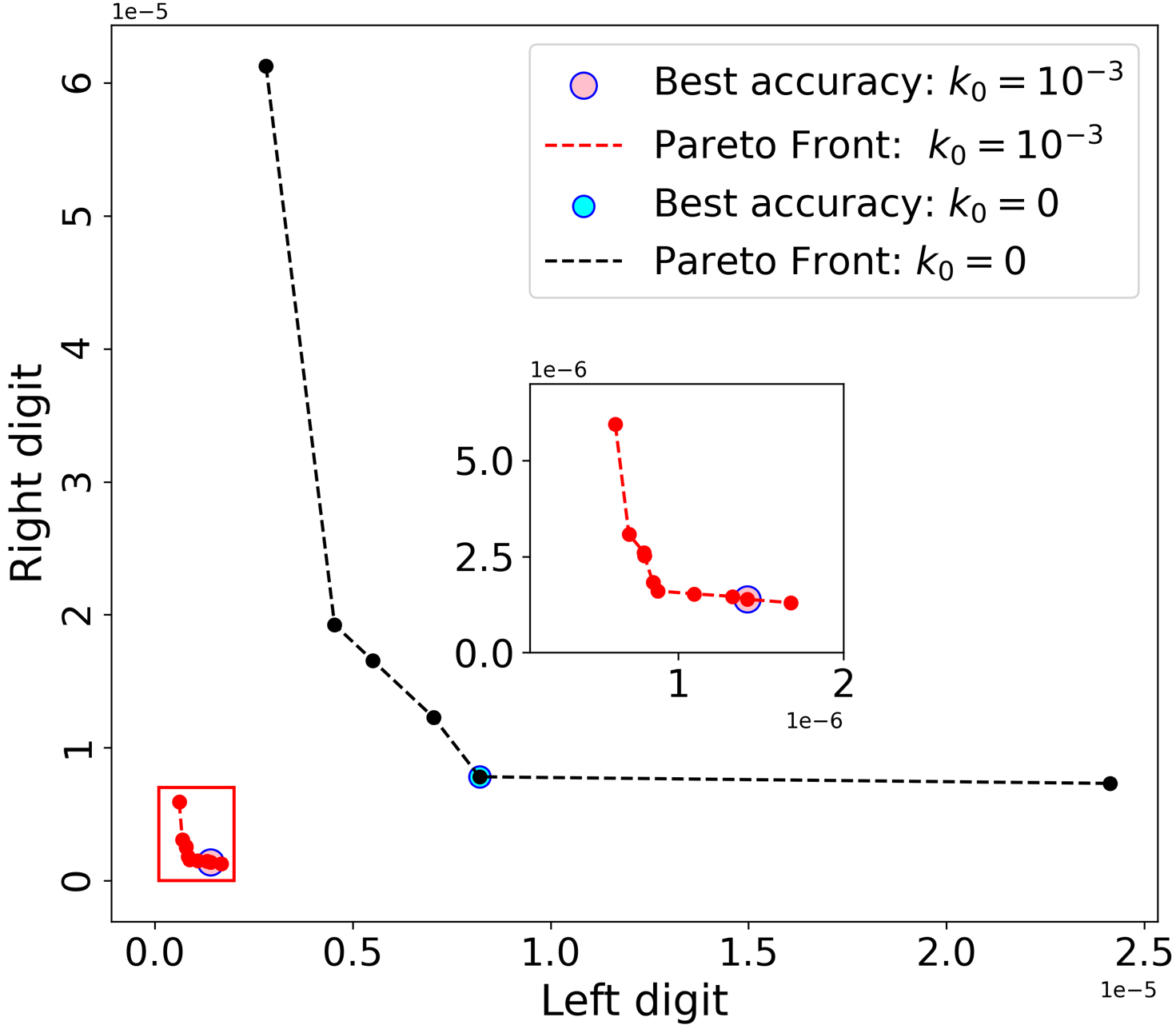}
        \label{fig:subfig1_1}
    \end{subfigure}%
    \begin{subfigure}[t]{0.245\textwidth}
        \caption{MDMTN on Cifar10Mnist}
        \includegraphics[width=1.0\textwidth]{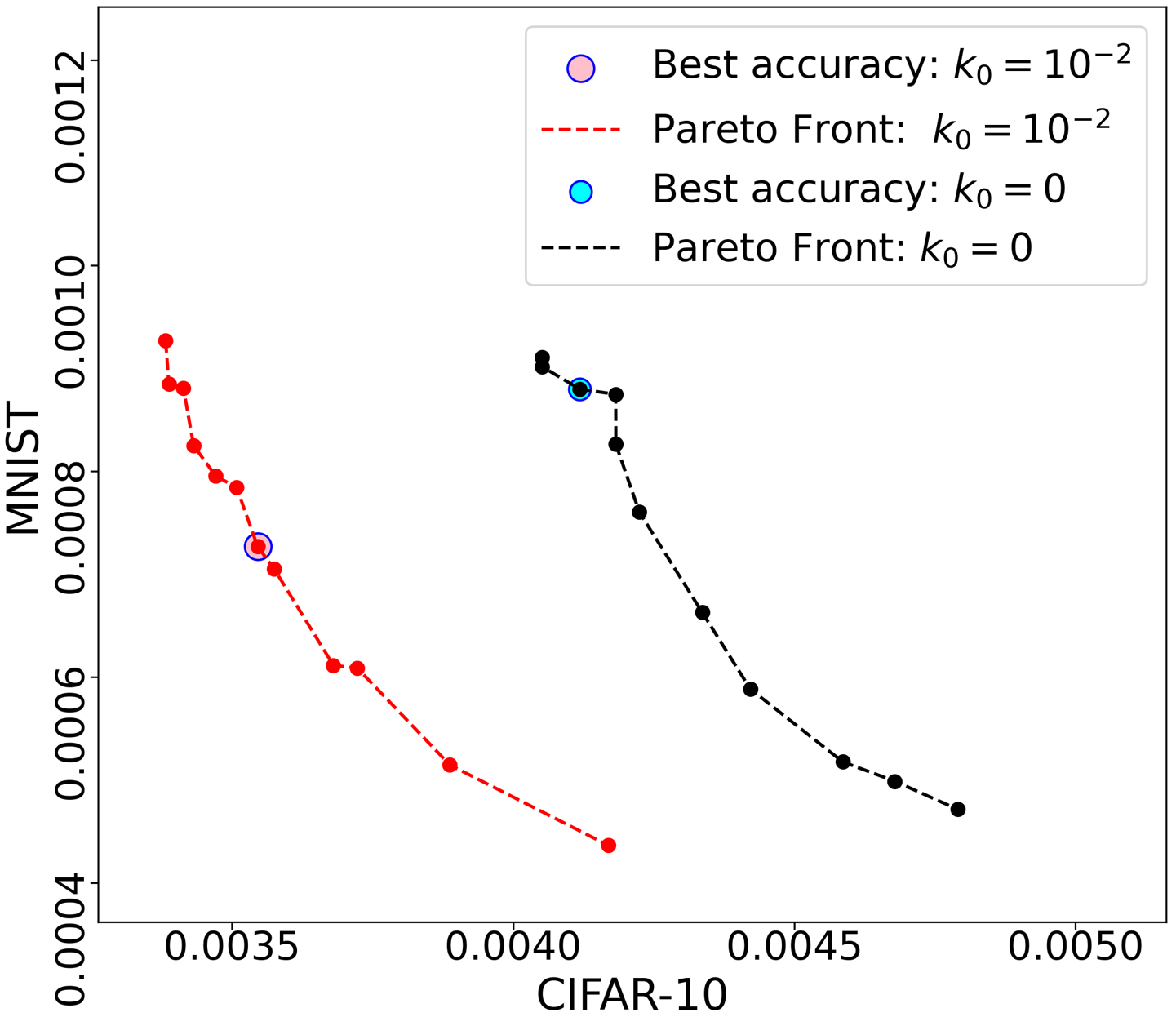}
        \label{fig:subfig1_1}
    \end{subfigure}%
    \hspace{0.01\textwidth}
\begin{subfigure}[t]{0.245\textwidth}
        \caption{HPS on MultiMNIST}
        \includegraphics[width=1.\textwidth]{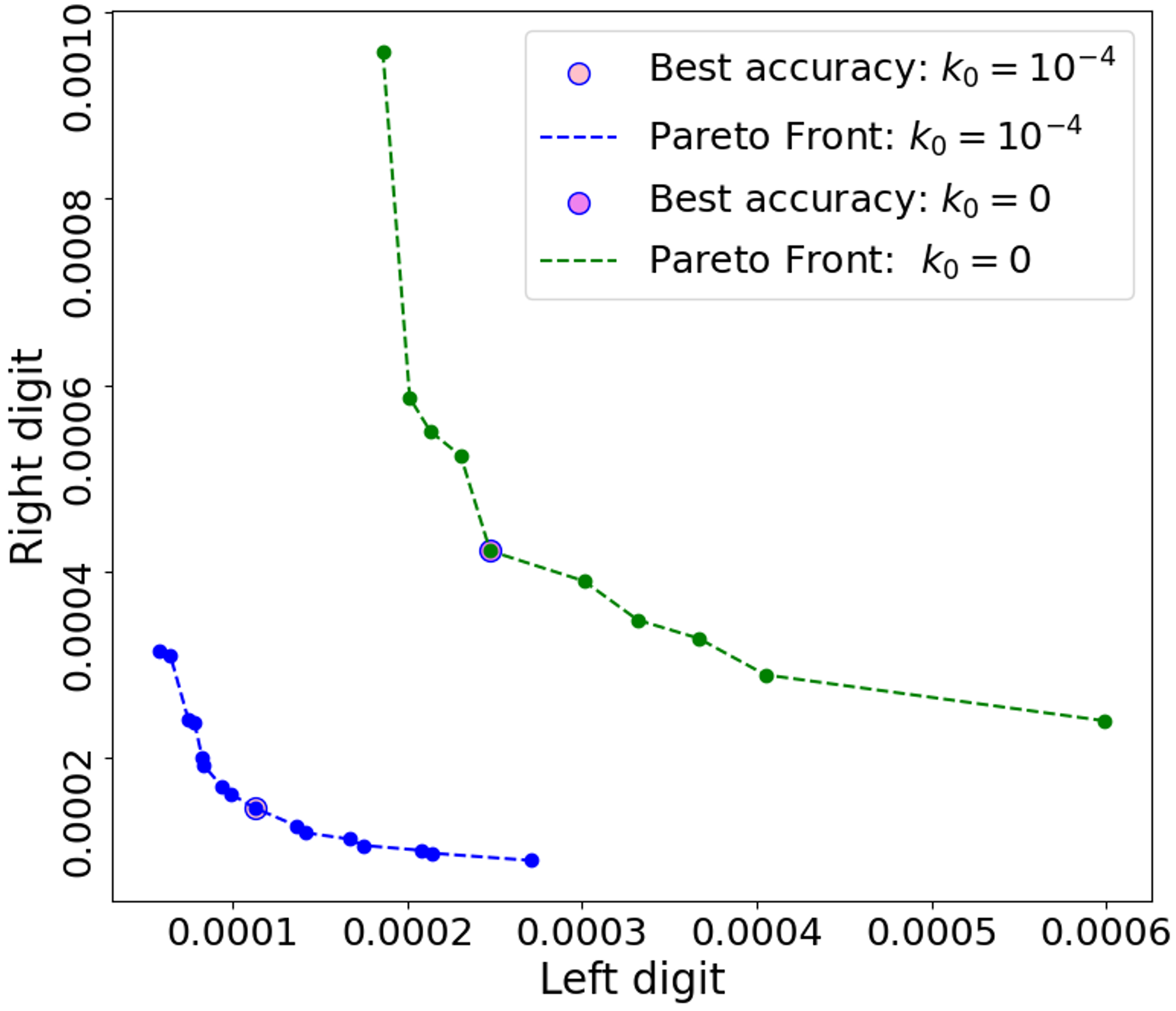}
        \label{fig:subfig1_2}
    \end{subfigure}%
    \begin{subfigure}[t]{0.245\textwidth}
        \caption{HPS on Cifar10Mnist}
        \includegraphics[width=1.\textwidth]{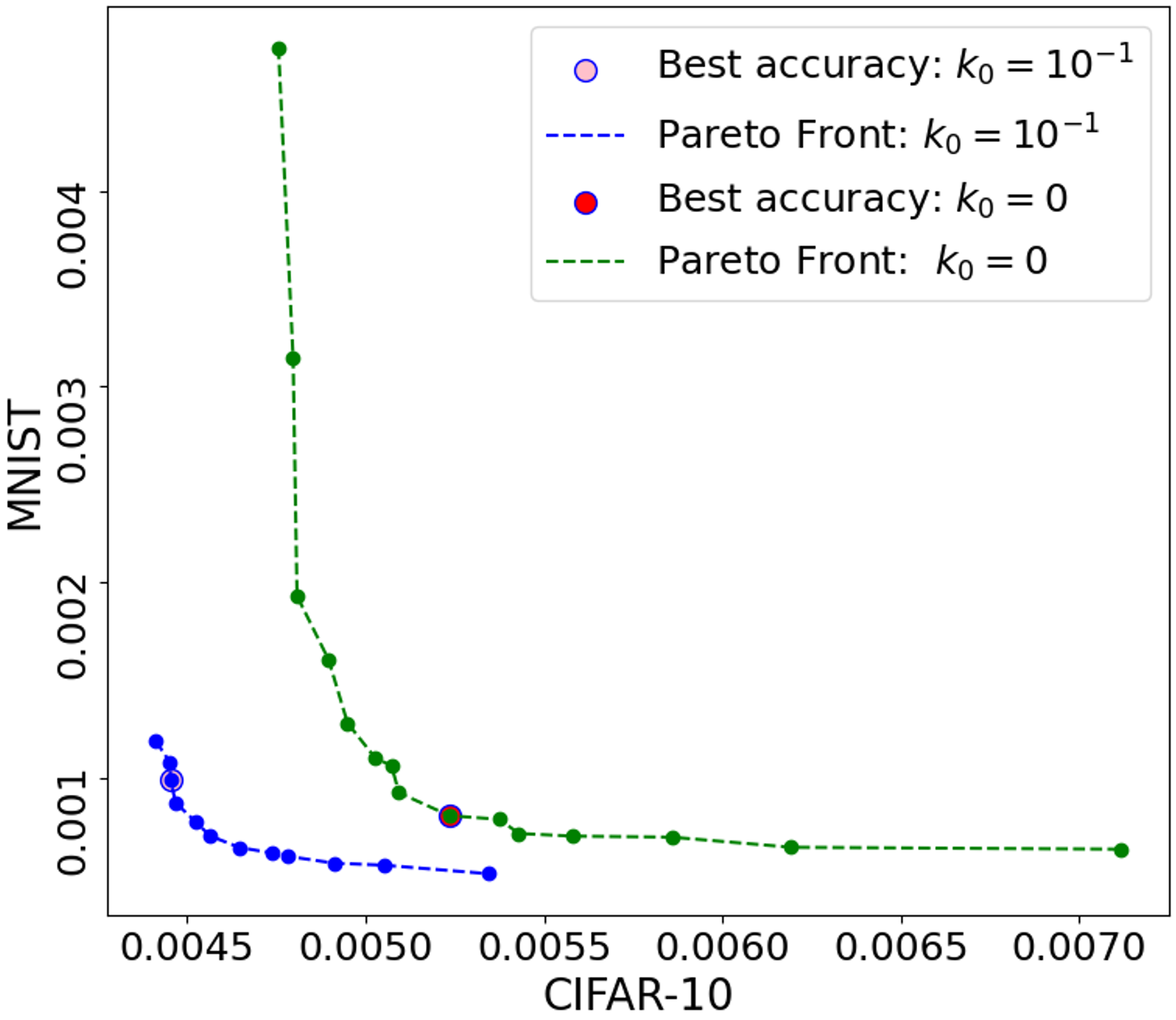}
        \label{fig:subfig1_2}
    \end{subfigure}%
    \hspace{0.05\textwidth}
\begin{subfigure}[t]{0.245\textwidth}
        \caption{Comparison on MultiMNIST}
        \includegraphics[width=1.05\textwidth]{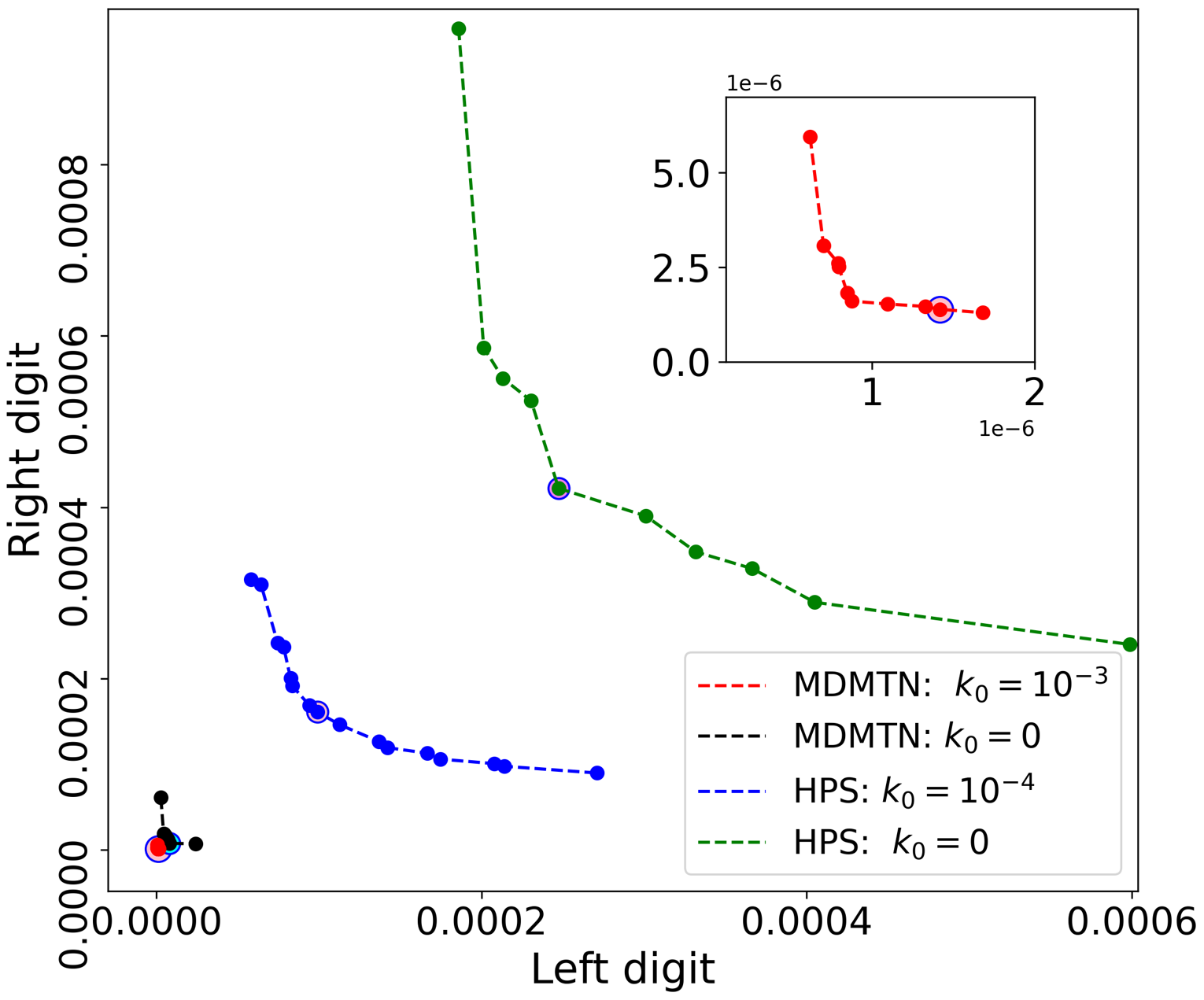}
        \label{fig:subfig1_3}
    \end{subfigure}%
\begin{subfigure}[t]{0.245\textwidth}
        \caption{Comparison on Cifar10Mnist}
        \includegraphics[width=1.\textwidth]{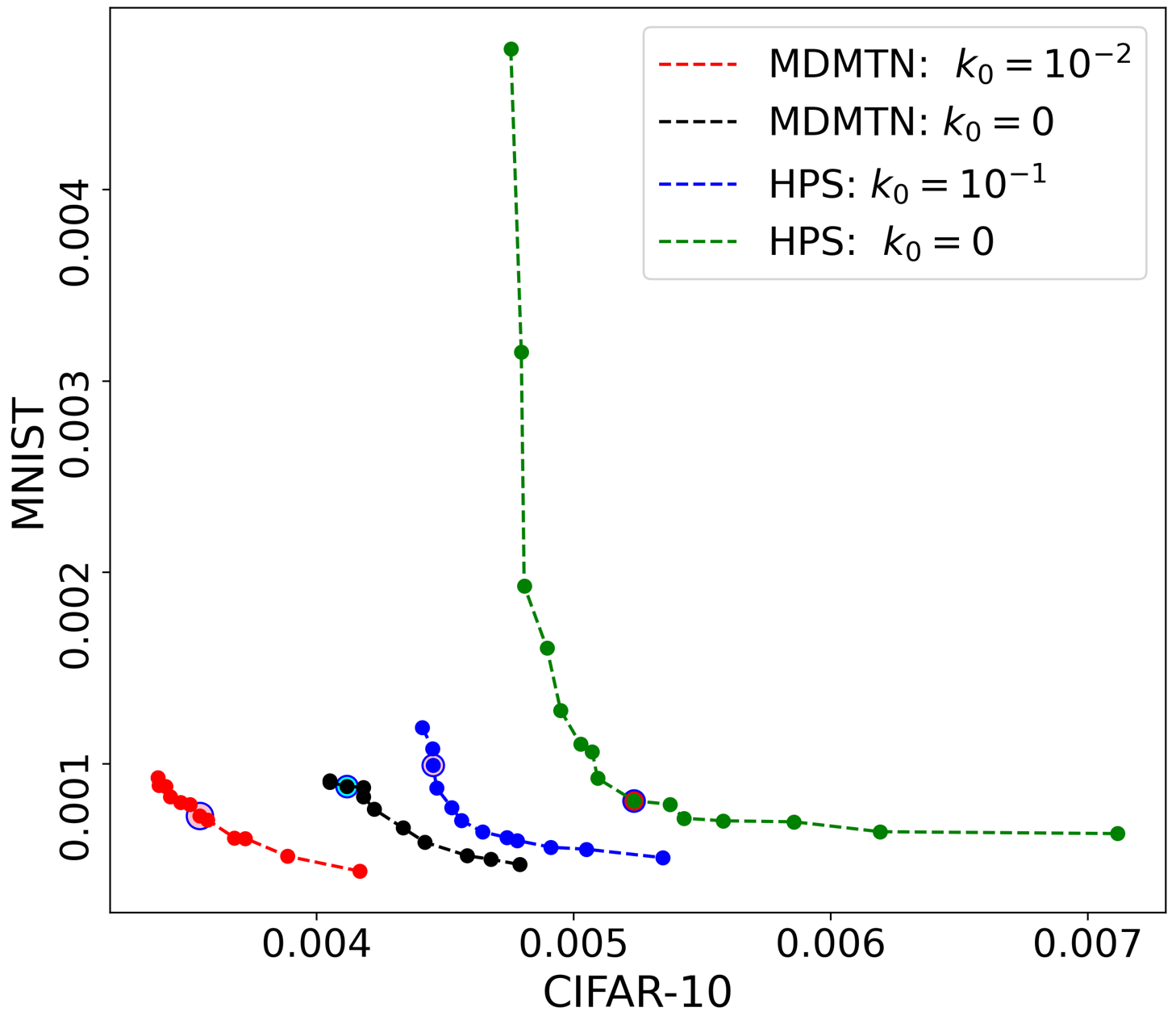}
        \label{fig:subfig1_3}
    \end{subfigure}%
    \caption{Comparison of 2D Pareto Fronts for the main tasks, with different model architectures}
    \label{fig:ComparePareto_MMCM}
\end{figure}

 When we discard the sparsity objective function, we get models with performance in the range of $75.69\% - 77.22\%$. 
Therefore, when dealing with significantly dissimilar tasks, the effect of sparsity on model performance could display randomness, although it holds the potential for improvement. We intend to further investigate this aspect in our upcoming research endeavors.

We report the training time for our method's retraining phase, KDMTL's main training phase, MGDA training repetitions, MTAN's training, and the combined training times of STL method for the main tasks.
Remarkably, our method typically demonstrates shorter training times compared to alternative approaches on both datasets, underscoring its efficiency and computational advantage.

Figure \ref{fig:ComparePareto_MMCM} shows how efficient a sparse model obtained with GrOWL is, compared to the original one, in generating the Pareto front for the main tasks. Our MDMTN model architecture excels at finding solutions that simultaneously optimize the main tasks while incorporating sparsity $(k_0 \neq 0)$. 
Therefore, MDMTN inherently possesses a stronger ability to capture task-related information while still minimizing unnecessary redundancy in the learned parameters. Furthermore, the fact that both MDMTN Pareto fronts $(k_0 \neq 0,\ \text{and}\ k_0 = 0)$ outperform those of HPS demonstrates that our model architecture excels at establishing an appropriate balance between primary task performance, regardless of whether sparsity is explicitly considered as an objective. 

\section{Conclusion}
This work presents a novel approach to addressing challenges arising from conflicting optimization criteria in various Deep Learning contexts. Our Multi-Objective Optimization technique enhances the efficiency and applicability of training Deep Neural Networks (DNNs) across multiple tasks by combining a modified Weighted Chebyshev scalarization with an Augmented Lagrangian method. Through empirical validation on both the MultiMNIST and our newly introduced Cifar10Mnist datasets, our technique, complemented by our innovative Multi-Task Learning model architecture named the Monitored Deep Multi-Task Network (MDMTN), has demonstrated superior performance compared to competing methods. These results underscore the potential of adaptive sparsification in training networks tailored to specific contexts, showcasing the feasibility of substantial network size reduction while maintaining satisfactory performance levels. Future research will delve into the integration of adaptivity mechanisms and explore strategies to mitigate the impact of sparsity on model performance in scenarios involving dissimilar tasks.

\section*{Acknowledgment}
This project received funding from the German Federal Ministry of Education and Research (BMBF) through the AI junior research group \enquote{Multicriteria Machine Learning}.

\bibliography{IEEE_references}
\bibliographystyle{IEEEtranS}

\newpage
\appendix

\subsection*{Proper Pareto Optimality:}
\begin{definition}
A solution $x^*$ of a MOP is said to be \textbf{properly Pareto Optimal} with respect to $\Omega$, in the sense of \cite{GEOFFRION1968618}, if there exists $M>0$ such that $\forall i \in \{0, ..., m\}\ \text{and}\ \forall x'\in \Omega$:

\resizebox{0.4\linewidth}{!}{$\big(\mathcal{L}_i(x') < \mathcal{L}_i(x^*)\big)\implies$}

        \resizebox{0.8\linewidth}{!}{$\bigg(\exists j \in \{0, ..., m\} \bigg| \\ \mathcal{L}_j(x^*) < \mathcal{L}_j(x')\ \text{,}\ \dfrac{\mathcal{L}_i(x^*) - \mathcal{L}_i(x')}{\mathcal{L}_j(x') - \mathcal{L}_j(x^*)} \le M\bigg)$}
\end{definition}

\begin{theorem}[\cite{kaliszewski2012quantitative}, pp. 48-50.]
\label{thm:nod_WCS_sol}
For any importance vector $k$, with $\ k_i \ge 0,\ \forall i\in\{0,...,m\}$, a  decision variable $x^* \in \Omega$ is an optimal solution of the Modified Weighted Chebyshev scalarization problem if and only if $x^*$ is properly Pareto optimal.
\end{theorem}

\subsection*{Datasets:}
In addition to the MultiMNIST dataset, we create Cifar10Mnist using CIFAR-10 and MNIST data sources. Since the CIFAR-10 training set consists of $50000$ images and the MNIST training set contains $60000$ digits, we pad the first $50000$ digits from MNIST over the CIFAR-10 images after making them slightly translucent. For the test set, we pad the $10000$ CIFAR-10 images over the $10000$ MNIST digits. For the experiments involving our method, we partitioned the training sets of both the MultiMNIST and Cifar10Mnist datasets into Train and Validation sets. Notably, the Validation set's data volume equaled that of the Test data.

\subsection*{ Additional experimentation details:} 

\begin{figure*}[t]
    \centering
    \begin{subfigure}[b]{0.35\textwidth}
        \includegraphics[width=1.1\linewidth]{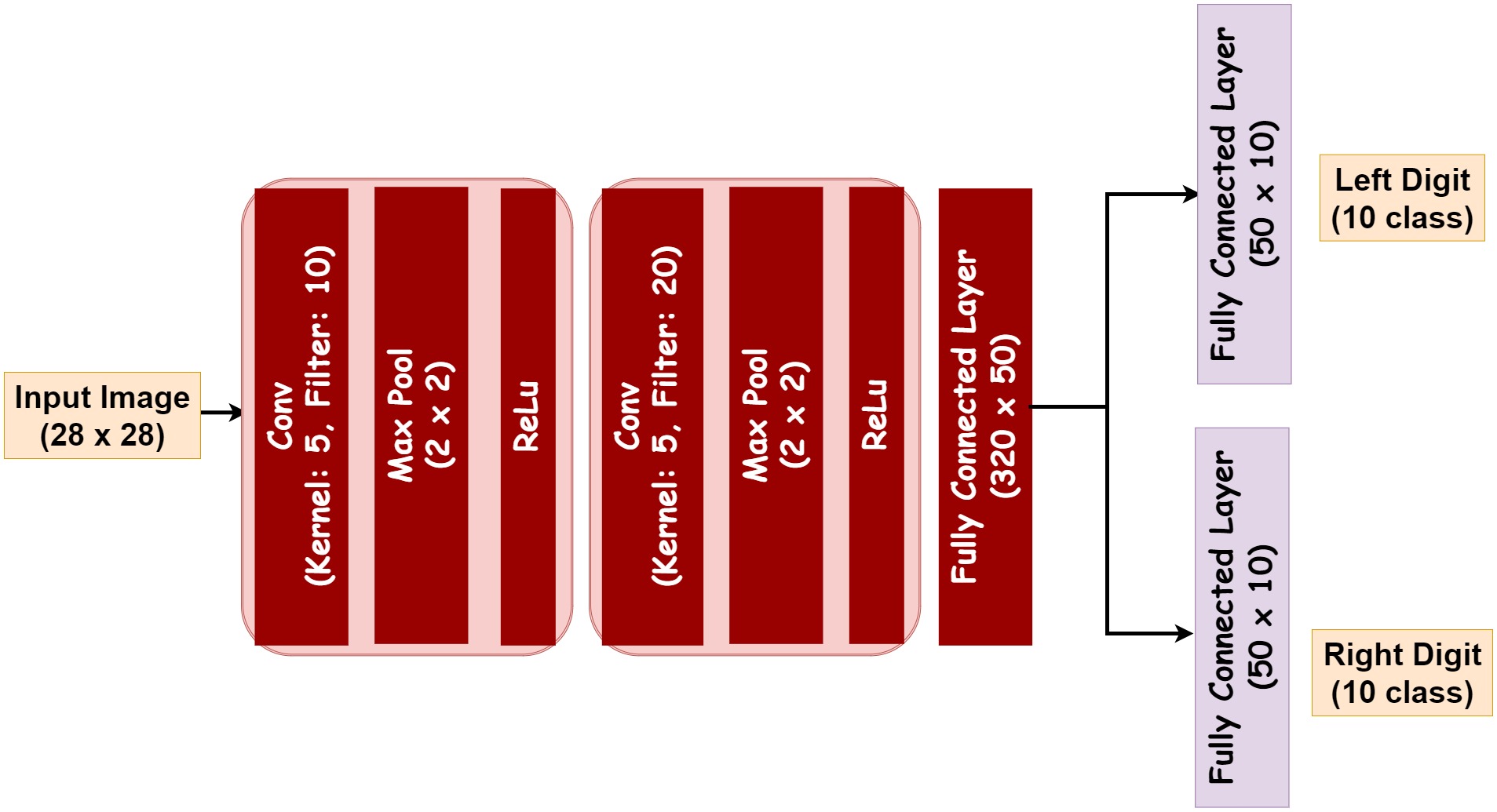}
        \caption{(MultiMNIST): HPS model}
        \label{subfig:subfigure2}
    \end{subfigure}
    \hspace{0.05\textwidth}
    \begin{subfigure}[b]{0.4\textwidth}
        \includegraphics[width=1.1\linewidth]{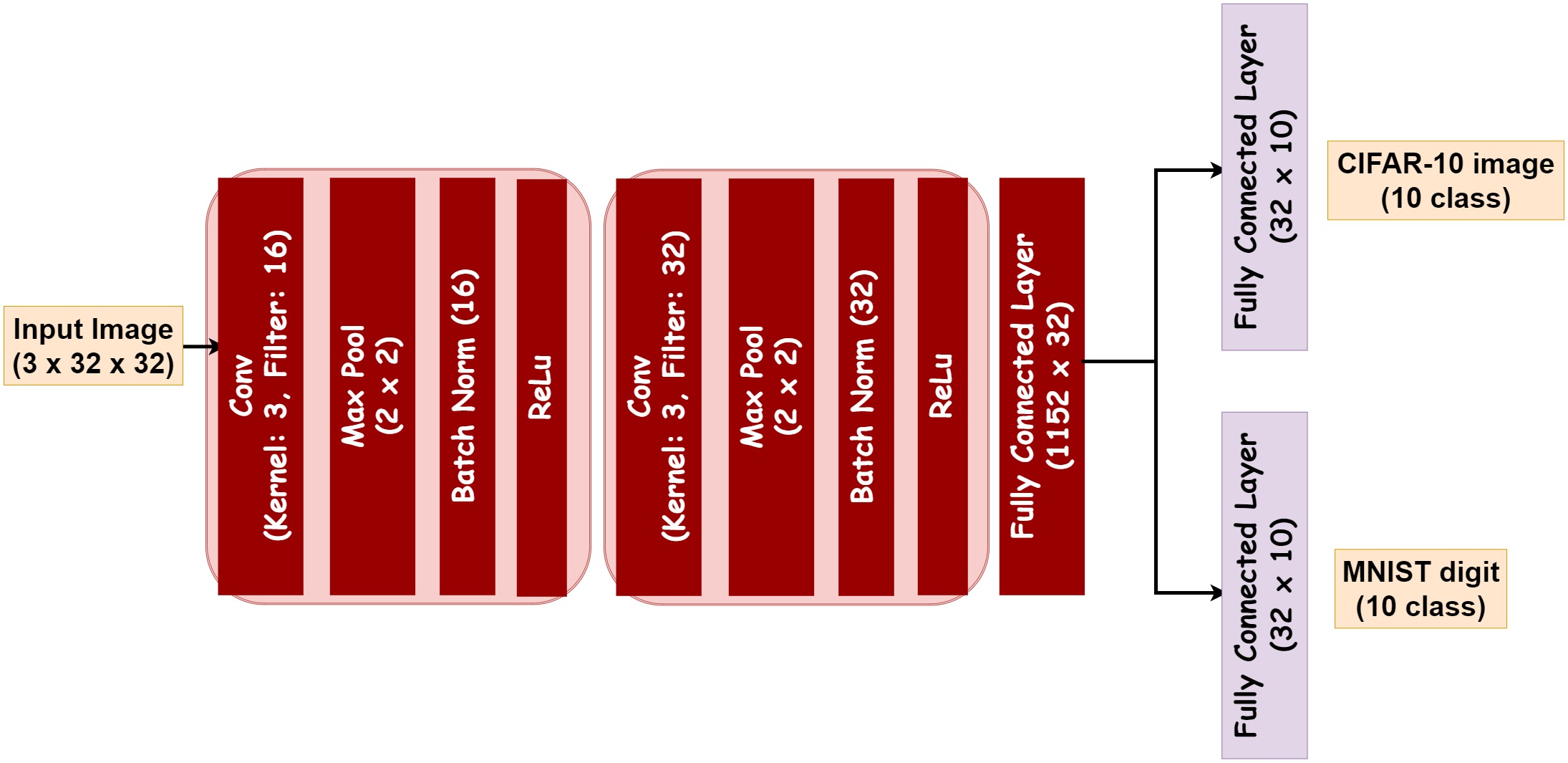}
        \caption{(Cifar10Mnist): HPS model}
        \label{subfig:subfigure2}
    \end{subfigure}
    \begin{subfigure}[b]{0.35\textwidth}
        \includegraphics[width=1.1\linewidth]{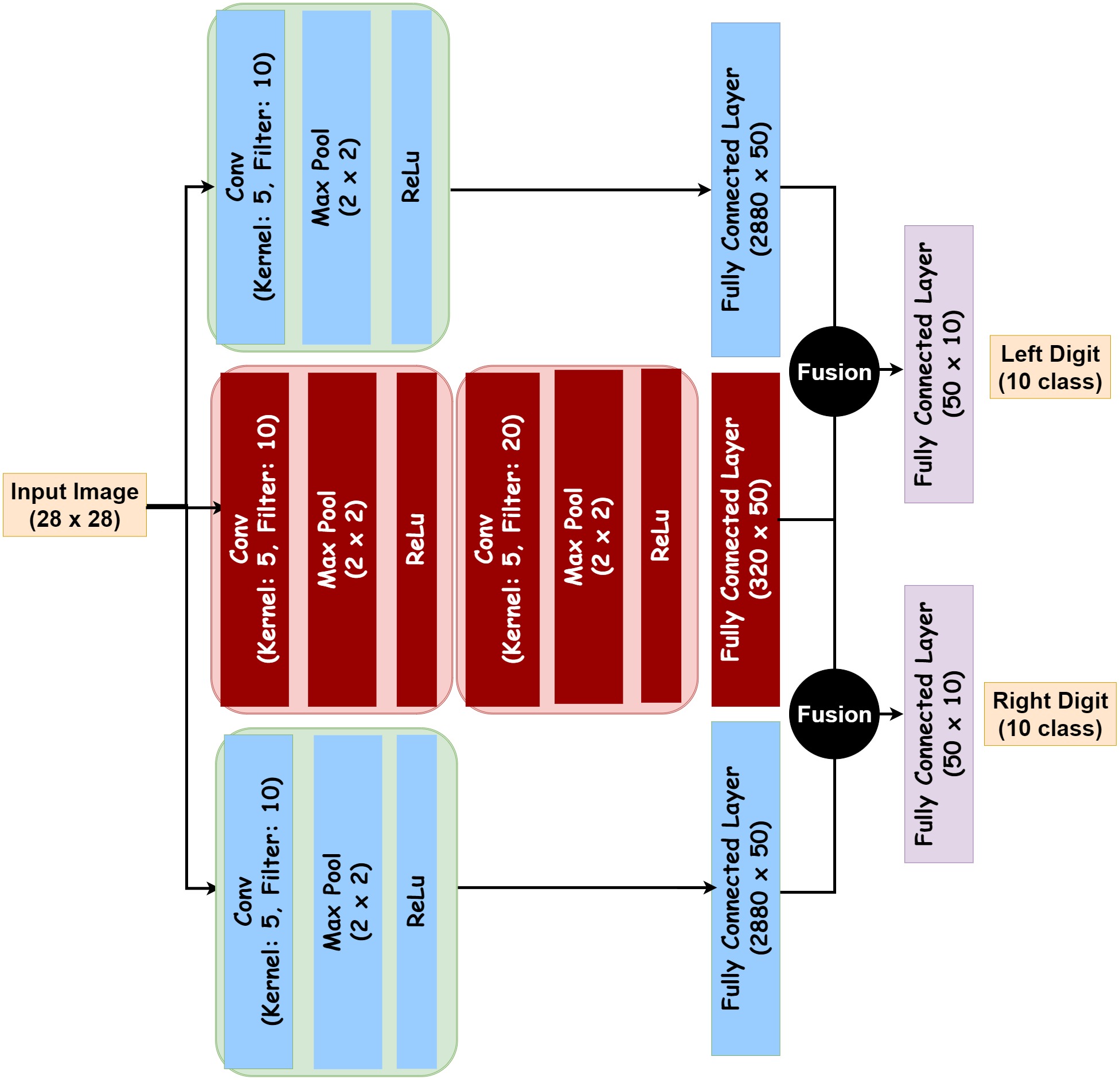}
        \caption{(MultiMNIST): MDMTN model}
        \label{subfig:subfigure3}
    \end{subfigure}
    \hspace{0.05\textwidth}
    \begin{subfigure}[b]{0.4\textwidth}
        \includegraphics[width=1.1\linewidth]{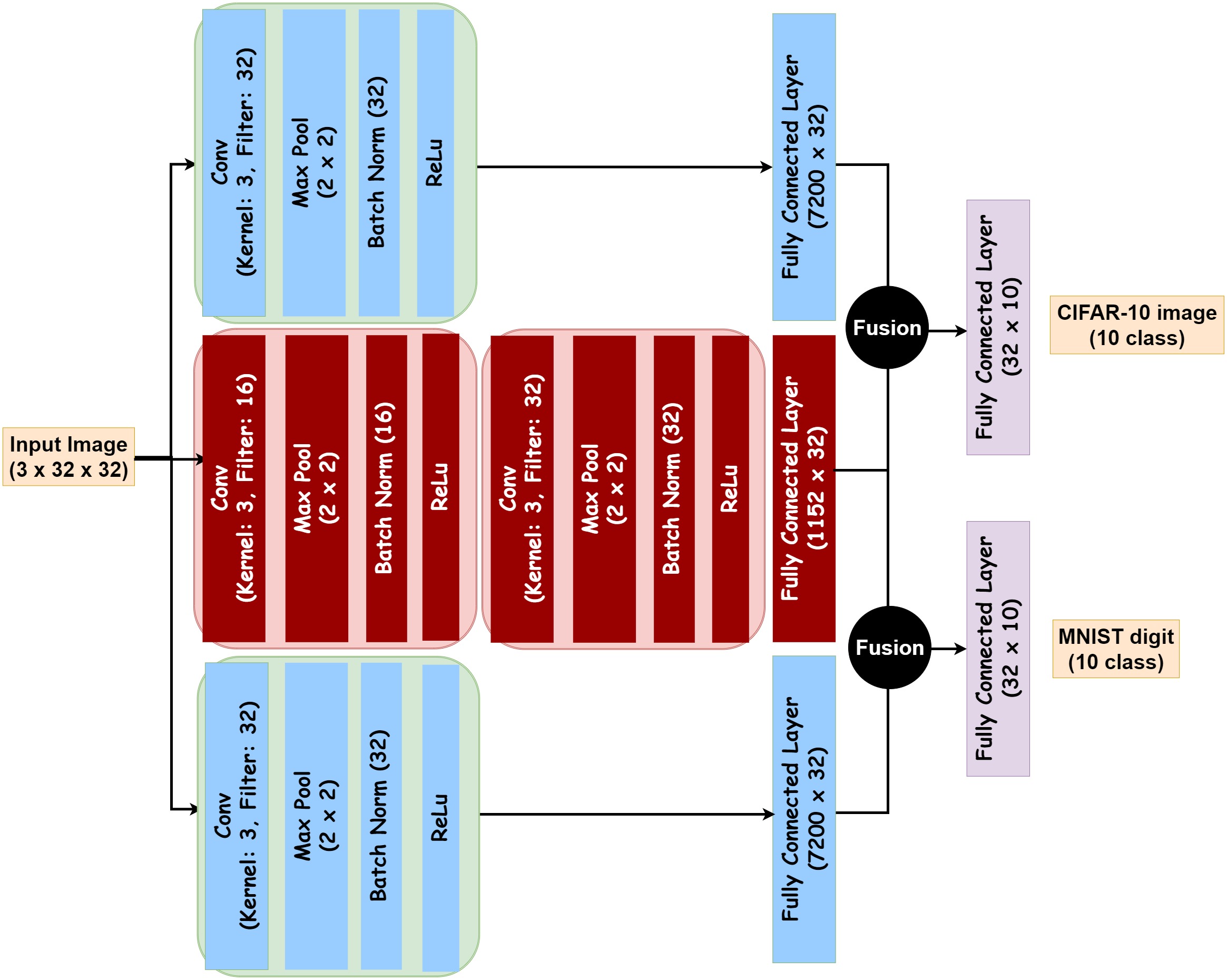}
        \caption{(Cifar10Mnist): MDMTN model}
        \label{subfig:subfigure3}
    \end{subfigure}
    \hspace{0.0\textwidth}
    \begin{subfigure}[b]{0.4\textwidth}
        \includegraphics[width=1.1\linewidth]{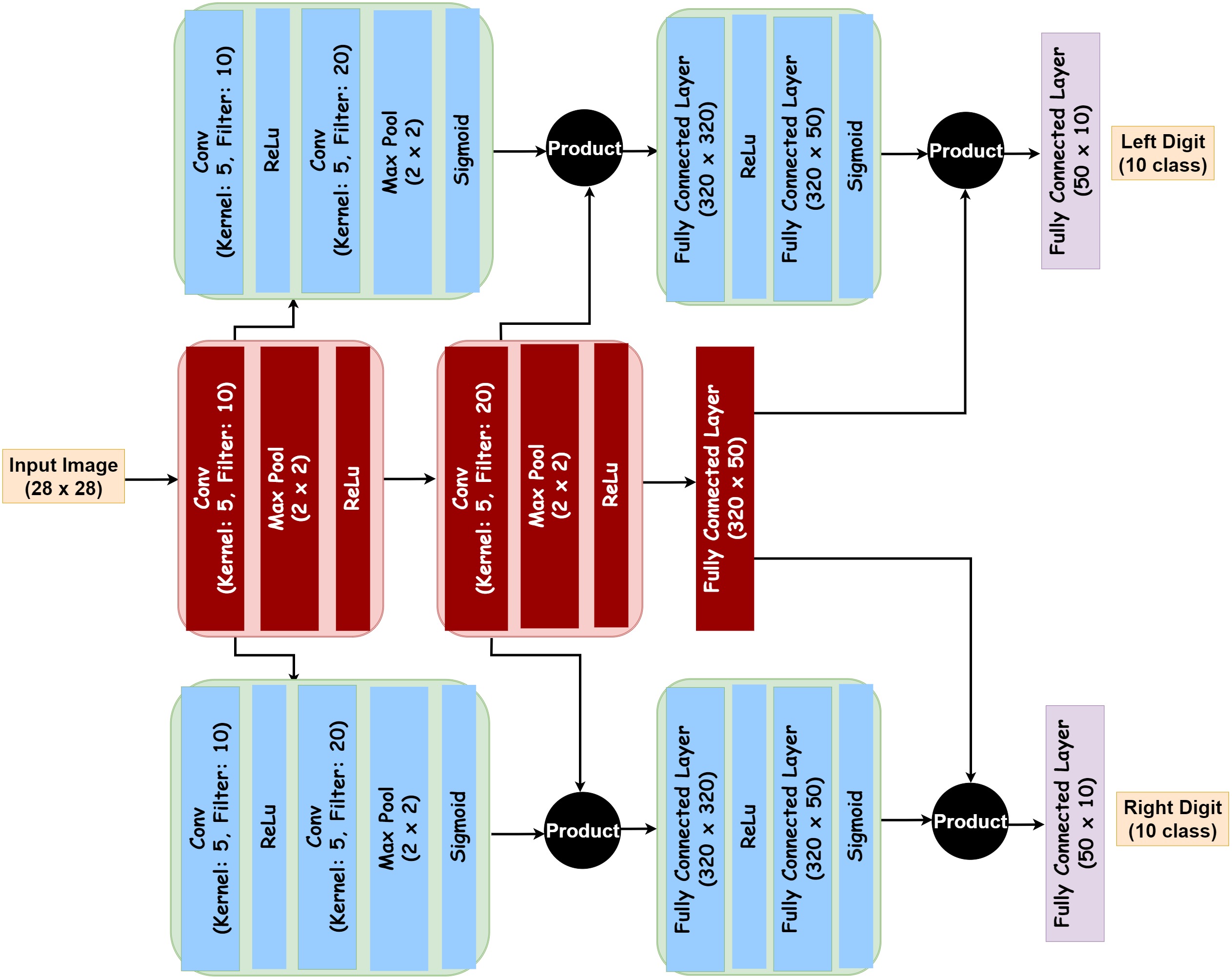}
        \caption{(MultiMNIST): MTAN model}
        \label{subfig:subfigure3}
    \end{subfigure}
    \hspace{0.05\textwidth}
    \begin{subfigure}[b]{0.4\textwidth}
        \includegraphics[width=1.1\linewidth]{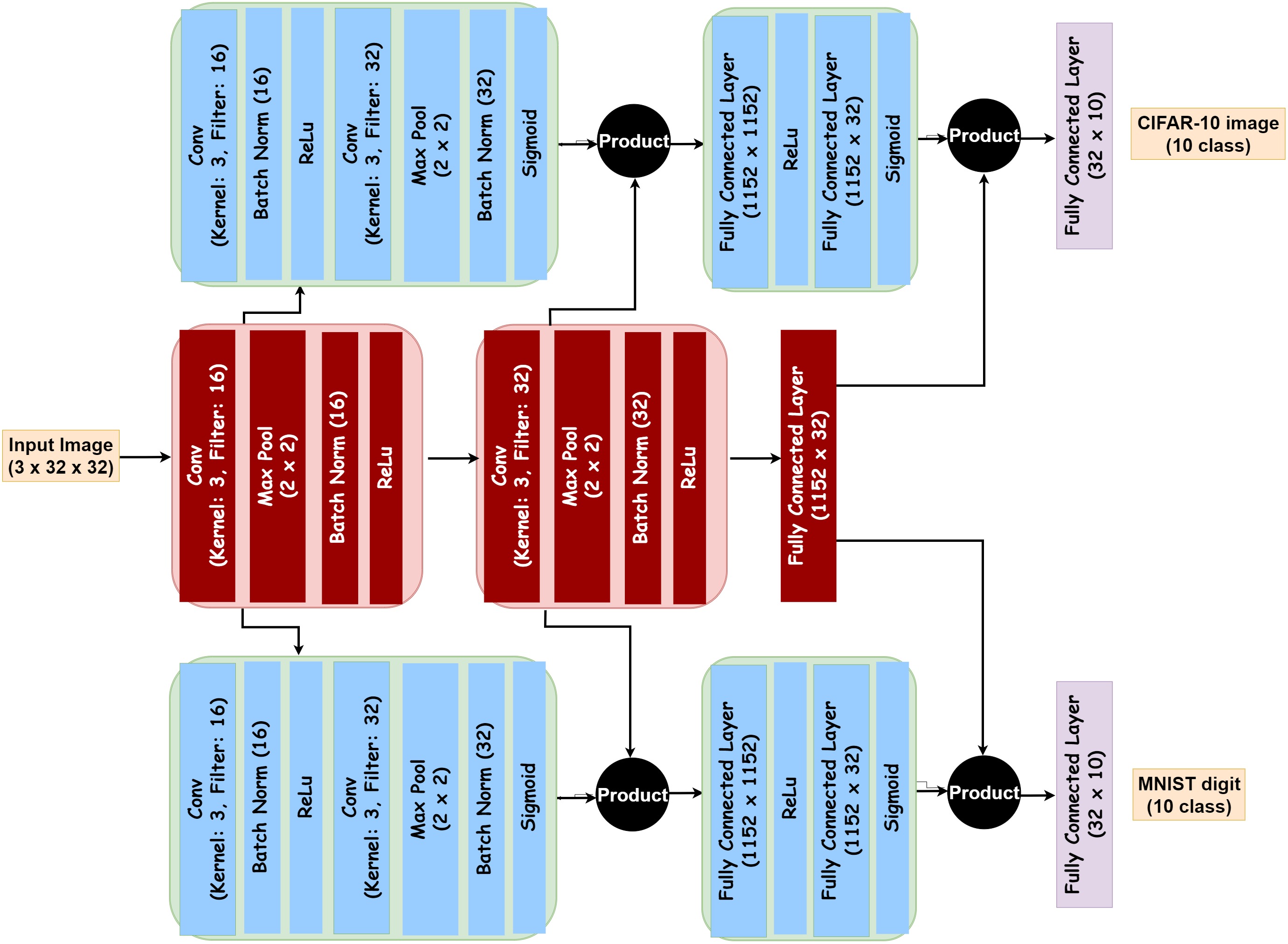}
        \caption{(Cifar10Mnist): MTAN model}
        \label{subfig:subfigure3}
    \end{subfigure}
    \caption{MTL models for MultiMNIST and Cifar10Mnist datasets.}
    \label{fig:MTLsMM}
\end{figure*}


Figure \ref{fig:MTLsMM} shows the HPS model considered for each dataset, as well as the corresponding MDMTN and MTAN model architectures. In each case, a Single Task-Learning (STL) model is the DNN model formed with the Shared network and the related Task-specific output network. For KDMTL architectures, the parameters of HPS networks are optimized to generate similar features with single-task networks. This involves using two fully connected layers \textit{(32 x 32)} and two fully connected layers \textit{(50 x 50)} as task-specific adaptors\cite{li2020knowledge} on Cifar10Mnist and MultiMNIST data, respectively.  We use PyTorch \cite{paszke2019pytorch} for all implementations and choose Adam \cite{kingma2014adam} as the optimizer. On our new dataset (Cifar10Mnist), we initialize the learning rate with $lr_s = 5\times 10^{-2}$ and train the STL models for 100 epochs. The learning rate is reduced by a factor of $0.98$ after each epoch.
For training the MTL models, we initialize the Lagrange multipliers with $\mu = 2.5\times 10^{-8}$ for all experiments on MultiMNIST and $\mu = 6.8\times 10^{-8}$ for those on Cifar10Mnist. After each iteration, we double the value of $\mu$\cite{bertsekas2014constrained}. The learning rate used for all MTL experiments with our method on MultiMNIST is $ lr_1 = 2.5\times 10^{-3}$, and $ lr_2 = 10^{-4}$ on Cifar10Mnist. We decrease $lr_1$ by a factor of $0.5$, and $lr_2$ by a factor of $0.98$ after each iteration. The first and second training stages are executed with $3$ and $10$ iterations, respectively, and $3$ epochs for each iteration. The sparsity rates $s_{min}$ and $s_{l_{max}}$ are respectively $20\%$ and $80\%$ for MDMTN models, and $10\%$ and $80\%$ for HPS models. On the Cifar10Mnist dataset, we use $10\%$ and $30\%$ for MDMTN models, and $5\%$ and $15\%$ for HPS models. For all experiments, we use $ 10^{-4}$ as $\epsilon$ value in the (Modified WC$(k, a)$) problem, and initialize the networks with regard to the Layer-Sequential Unit-Variance Initialization (LSUV) approach \cite{mishkin2015all}.  The similarity preference used in the \textit{Affinity propagation} method for all experiments on MultiMNIST is 0.7, and 0.8 for those on Cifar10Mnist.

When evaluating the MGDA method on our proposed MDMTN model architecture, we follow \cite{sener2018multi} and use SGD as the optimizer. We train the models
for $100$ epochs and repeat the training $10$ times on MultiMNIST
and $5$ times on Cifar10Mnist. The learning rate is initialized with $10^{-2}$ and is halved after a period of 30 epochs, while the momentum is set to $0.9$. 

 We conduct hyperparameter tuning for both KDMTL and MTAN methods, specifically optimizing the weighted parameters of the main tasks and the learning rate using a subset of the training data. 


\begin{figure*}[t]
    \centering
    \begin{subfigure}[b]{0.45\textwidth}
        \includegraphics[width=1\linewidth]{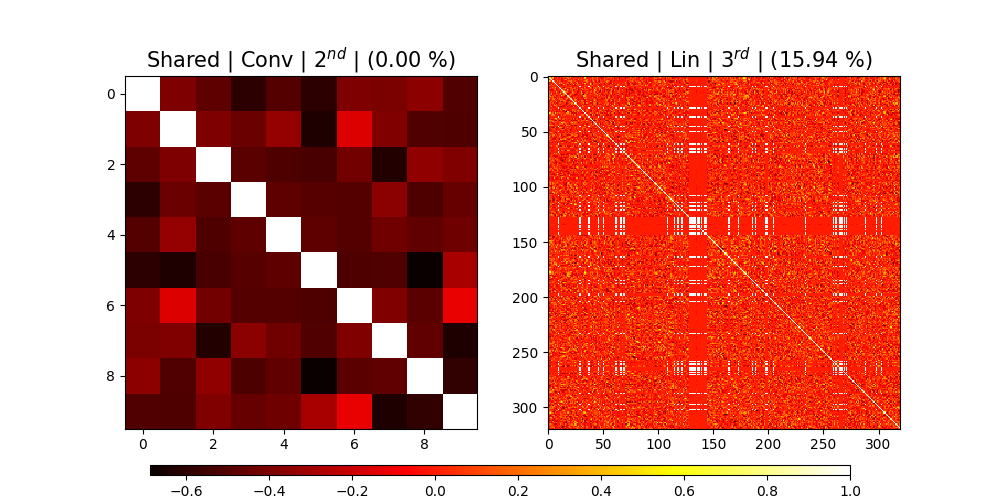}
        \caption{(MultiMNIST): HPS layers}
        \label{subfig:subfigure2}
    \end{subfigure}
    \begin{subfigure}[b]{0.45\textwidth}
        \includegraphics[width=1\linewidth]{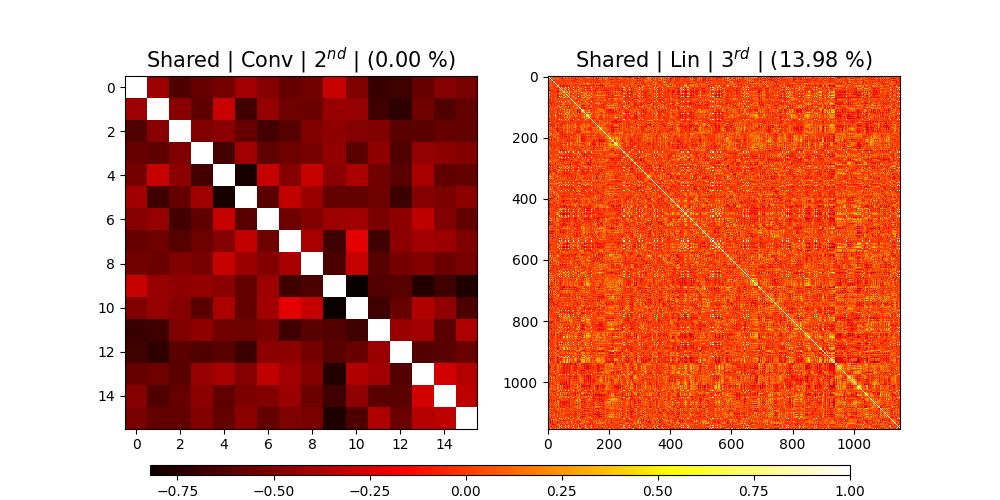}
        \caption{(Cifar10Mnist): HPS layers}
        \label{subfig:subfigure2}
    \end{subfigure}
    \begin{subfigure}[b]{0.9\textwidth}
        \includegraphics[width=1\linewidth]{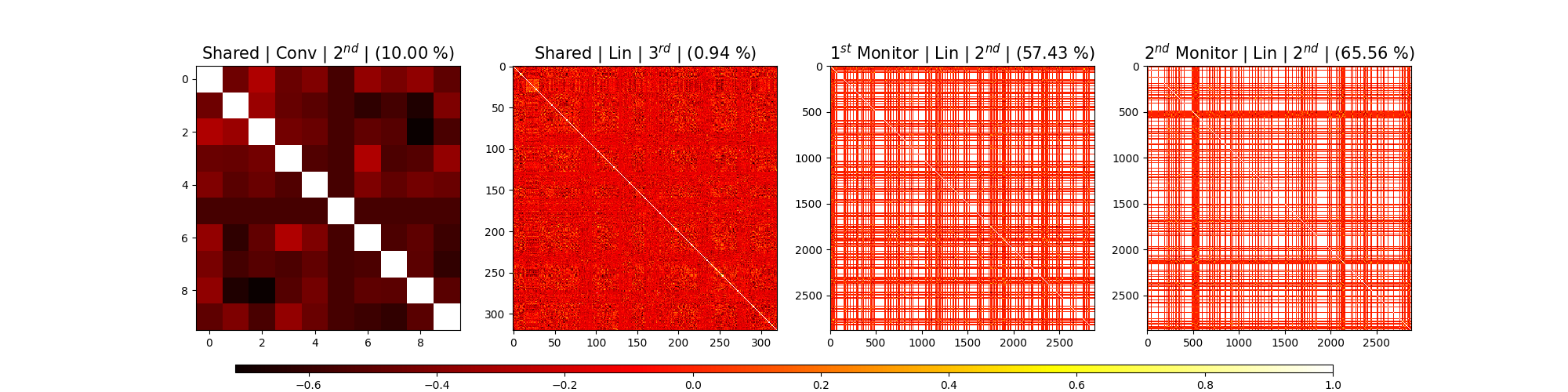}
        \caption{(MultiMNIST): MDMTN layers}
        \label{subfig:subfigure3}
    \end{subfigure}
    \begin{subfigure}[b]{0.9\textwidth}
        \includegraphics[width=1.\linewidth]{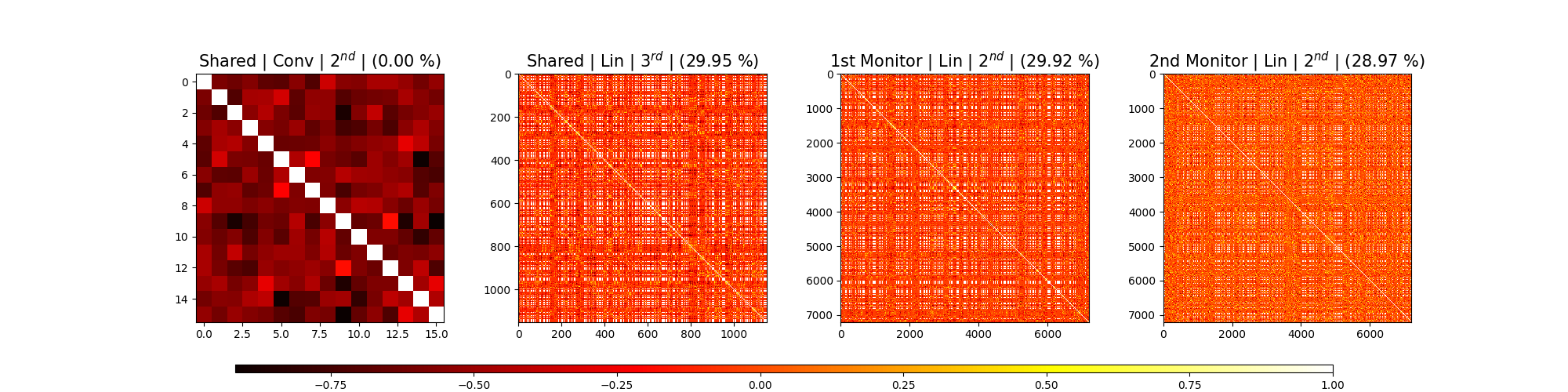}
        \caption{(Cifar10Mnist): MDMTN layers}
        \label{subfig:subfigure3}
    \end{subfigure}
    \caption{Pairwise similarity maps of the outputs (rows of the weight matrices) from layers that undergo GrOWL regularization. The black and white areas indicate perfect similarity (negative and positive, respectively) between the related outputs. Each map is titled as follows: (Shared Network or Monitor) $|$ (Convolutional or Linear) $|$ (Layer Position) $|$ (Sparsity Rate)
}
    \label{fig:MTLsSimilarityMaps}
\end{figure*}


\subsection*{Additional Results:}
For each preference vector, we got a suitable sparse model from the same original model. To generate a 2D Pareto front for the main objective functions (Figure \ref{fig:ComparePareto_MMCM}), we now consider the sparse model architecture with the highest performance on the primary tasks, as well as its associated sparsity coefficient $k_0$. This model is trained through a standard single training stage (without forcing parameter sharing or additional sparsification). On the MultiMNIST dataset, this boosts the global performance of the model to $97.075\%$ of accuracy ($97.43\%$ for the left digits and $96.72\%$ for the right digits) for $k = (10^{-3}, 0.199, 0.8 )$. For the Cifar10Mnist dataset, the global performance of the model reaches $78.354\%$ of accuracy ($61.71\%$ for CIFAR-10 images and $95.0\%$ for MNIST digits) with $k = (10^{-2}, 0.4, 0.59 )$.  
Since the preference vector that outperformed others during this Pareto front study is not the one that produced the sparse model used, we do not consider these performances in our main results (Table 1).
So, the idea is mainly to assess how efficient a sparse model obtained with GrOWL may be, compared to the original one, in generating the Pareto Front for the main tasks. That's why we also ignore parameter sharing here. 
Excluding the MGDA and KDMTL methods from this supplementary study was deliberate, as they rely on distinct forms of loss function formulations unique to each method. Moreover, these methods only yield a singular point on the Pareto Front.

To evaluate the impact of applying the GrOWL function to layers during training, in relation to sparsity and parameter sharing, we present the similarity maps for those layers within each architecture in Figure \ref{fig:MTLsSimilarityMaps}. It becomes evident that certain parameters (neurons) within these layers exhibit a remarkable ability to entirely share their outputs. Furthermore, given that the sparsity rate of the Task Specific network layers is higher than that of the shared component overall, it is clear that the models continue to place greater reliance on the Shared network in comparison to the task-specific components. This upholds the essence of Multi-Task Learning, preventing the exclusion of shared representations across all tasks.

\end{document}